\documentclass{article} 
\usepackage{iclr2025_conference,times}
\usepackage{graphicx}
\usepackage{pgfplots}
\usepackage{multirow}
\usepackage{caption}
\usepackage{booktabs}
\usepackage{wrapfig}

\usepackage{amsmath,amsfonts,bm}

\def\eqref#1{equation~\ref{#1}}

\def\1{\bm{1}}

\DeclareMathAlphabet{\mathsfit}{\encodingdefault}{\sfdefault}{m}{sl}
\SetMathAlphabet{\mathsfit}{bold}{\encodingdefault}{\sfdefault}{bx}{n}

\usepackage{hyperref}
\usepackage{url}

\title{Verbalized Graph Representation Learning:\\
A Fully Interpretable Graph Model Based on Large Language Models \\ Throughout the Entire Process
}

\author{Xingyu~Ji$^{\ddagger}$, JiaLe~Liu$^{\ddagger}$, Lu~Li, Maojun~Wang \& Zeyu~Zhang\thanks{Corresponding author. $^{\ddagger}$ Xingyu Ji and Jiale Liu contributed equally to this work.} \\
Huazhong Agricultural University\\
Hubei, China \\
\texttt{zhangzeyu@mail.hzau.edu.cn} \\
}

\usepackage{amsthm}
\newtheorem*{theorem}{Theorem}

\iclrfinalcopy
\begin{document}
\maketitle

\begin{abstract}
Representation learning on text-attributed graphs (TAGs) has attracted significant interest due to its wide-ranging real-world applications, particularly through Graph Neural Networks (GNNs). Traditional GNN methods focus on encoding the structural information of graphs, often using shallow text embeddings for node or edge attributes. This limits the model to understand the rich semantic information in the data and its reasoning ability for complex downstream tasks, while also lacking interpretability. With the rise of large language models (LLMs), an increasing number of studies are combining them with GNNs for graph representation learning and downstream tasks. While these approaches effectively leverage the rich semantic information in TAGs datasets, their main drawback is that they are only partially interpretable, which limits their application in critical fields. In this paper, we propose a verbalized graph representation learning (VGRL) method which is fully interpretable. In contrast to traditional graph machine learning models, which are usually optimized within a continuous parameter space, VGRL constrains this parameter space to be text description which ensures complete interpretability throughout the entire process, making it easier for users to understand and trust the decisions of the model. We conduct several studies to empirically evaluate the effectiveness of VGRL and we believe this method can serve as a stepping stone in graph representation learning. The source code of our model is available at \url{https://anonymous.4open.science/r/VGRL-7E1E}
\end{abstract}

\section{Introduction}
Many real-world graphs incorporate textual data, forming what are known as Text-Attributed Graphs (TAGs)~\citep{yang2021graphformers}. In TAGs, nodes represent textual entitities such as papers, while edges denote relationships between them, such as citations or co-authorships. For instance, the Cora dataset can be modeled as a TAG, where each node represents a research paper, and the node attributes include features such as the paper’s title, abstract, and keywords. By integrating textual attributes with graph topology, TAGs facilitate more effective representation learning, making them valuable for tasks like document classification, clustering ~\citep{wang2023user}, citation analysis, and recommendation systems ~\citep{zhu2021textgnn, zhang2023contrastive}. This combination of textual and relational data offers deeper insights, especially when both content and connections are essential to the analysis. 

Although traditional Graph Neural Network (GNN) models, such as Graph Convolutional Network (GCN) \citep{kipf2016semi} and Graph Attention Network (GAT) ~\citep{velivckovic2017graph}, have achieved significant performance improvements across multiple tasks, they generally suffer from a lack of interpretability. As these models largely rely on complex network architectures and implicit feature learning processes, understanding their internal decision mechanisms and how specific features influence task outcomes becomes challenging, thereby limiting their transparency and trustworthiness in practical applications. To address this issue, researchers have proposed several interpretable GNN models.
These interpretable methods can generally be divided into input interpretability, training process interpretability, and decision-making process interpretability. For example, GNNExplainer ~\citep{ying2019gnnexplainer} is a method for input interpretability, which selects a small subgraph of the input graph together with a small subset of node features that are most influential for the prediction as an explanation, XGNN ~\citep{yuan2020xgnn} is a method for training process interpretability which reveals the basis of the model's predictions by generating interpretable graph structures, and SE-SGformer ~\citep{li2024se} is a method for decision-making process interpretability which constructs a novel explainable decision process by discovering the $K$-nearest (farthest) positive (negative) neighbors of a node for predicting edge signs. Clearly, while these methods all have a certain degree of interpretability, they can only explain one part of the entire process of model input, training, and output. Therefore, our goal is to  implement comprehensive interpretability by simultaneously achieving input interpretability, training process interpretability, and decision-making process interpretability.


\begin{figure}
    \centering
    \includegraphics[width=0.9\linewidth]{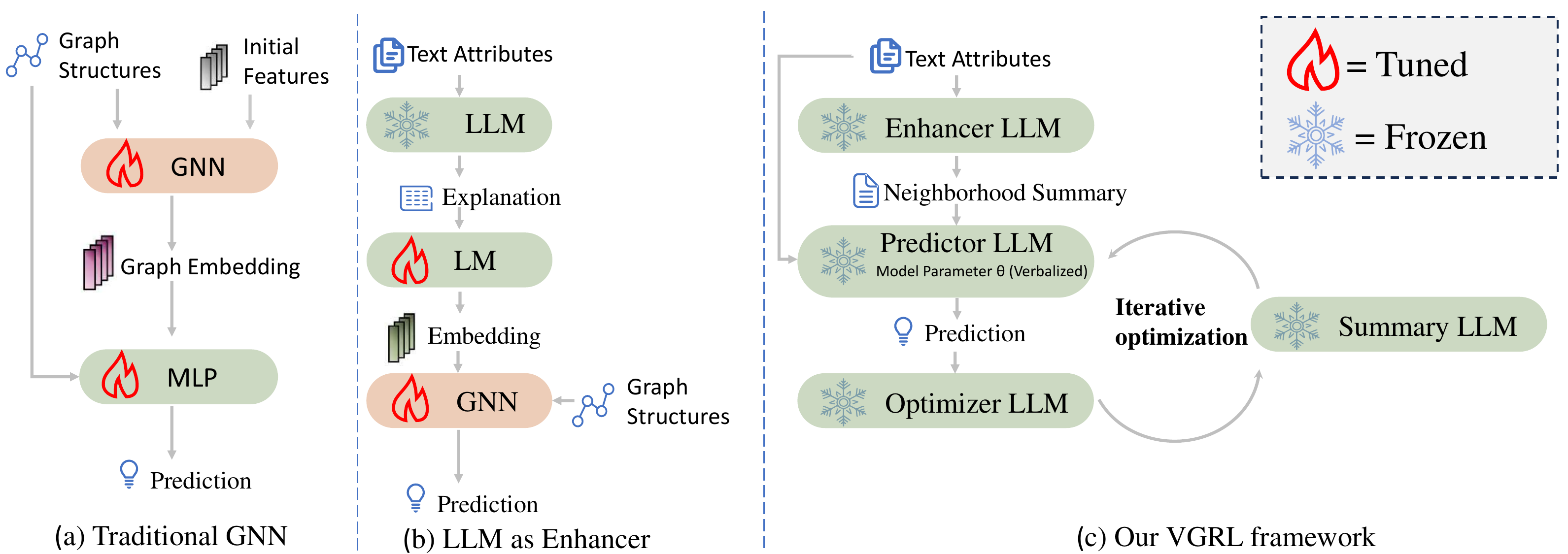}
    \caption{Comparison of Graph Representation Learning Methods
\textbf{(a)} Traditional Graph Neural Networks (GNNs) rely on graph structures and initial features for embedding generation and prediction.
\textbf{(b)} Incorporating a Language Model (LM) enhances GNNs, where a Large Language Model (LLM) provides explanations that refine the embedding process for improved predictions.
\textbf{(c)} Our proposed Verbalized Graph Representation Learning (VGRL) framework introduces an iterative optimization process involving multiple frozen LLMs (Enhancer, Predictor, Optimizer, and Summary), emphasizing interpretability and parameter tuning through verbalized model adjustments.}
    \label{fig:enter-label}
\end{figure}

In recent years, with the breakthroughs of large language models (LLMs) in the field of natural language processing, researchers have gradually begun to integrate them with GNNs to enhance model performance and capabilities.
For instance, LLMs can act as predictors ~\citep{tang2024graphgpt}, generating more accurate predictions by analyzing node features and structural information for the TAGs. Also, TAPE ~\citep{he2023harnessing} prompts a powerful LLM to explain its predictions and serve explanations as supplementary text attributes for the downstream LMs and GNN models. Due to the powerful text inference capabilities of LLMs, they are capable of processing TAGs, reasoning about the node classification prediction process of TAGs, and generating explanations in text that is comprehensible to humans. Therefore, we consider the use of LLMs to achieve comprehensive interpretability.
However, using LLMs to handle graph tasks and provide interpretability is not easy. Specifically, there are currently two main approaches to applying LLMs in the field of graph: one is to pre-train or fine-tune LLMs to adapt to various graph downstream tasks. But due to the vast number of parameters typically found in LLMs, the cost of fine-tuning LLMs is quite high and the training time is long. The second is to directly freeze the LLMs for inference but this method does not yield good results. For example, we directly froze the predictor LLMs for node classification prediction in subsequent experiments, and the prediction accuracy was generally not high, as shown in Table \ref{table:main}. 

In summary, we face two major challenges to achieve comprehensive interpretability with LLMs:

\textbf{Challenge 1:} How can we ensure that a model is interpretable in terms of input, training process, and decision-making simultaneously? 

\textbf{Challenge 2:} How can we optimize the performance of LLMs without fine-tuning the model parameters to reduce costs?

To address these challenges, we propose the \textbf{V}erbalized \textbf{G}raph \textbf{R}epresentation \textbf{L}earning (VGRL) method. \textbf{For Challenge 1}, VGRL utilizes a verbalized approach to create intuitive connections between input features and predictions and VGRL generates textual explanations at each iteration stage, helping researchers and practitioners better grasp the training dynamics of the model. Also, VGRL provides natural language descriptions for the model’s predictions, clearly explaining the rationale behind each decision. \textbf{For Challenge 2}, instead of relying on costly fine-tuning of the LLM parameters, VGRL leverages a prompt-based optimization strategy. This involves crafting task-specific prompts to guide the LLM in generating optimal predictions without modifying its internal parameters. By utilizing prompt engineering techniques, VGRL maintains high performance while significantly reducing computational costs associated with traditional fine-tuning methods. Additionally, this approach allows the model to remain versatile across various tasks, as it can be adapted to new datasets or problems simply by adjusting the prompts, further enhancing its efficiency and scalability.


Our contributions are as follows:
\begin{itemize}
    \item We propose a novel verbalized graph learning framework that ensures complete interpretability throughout the entire process, from input to training and decision-making, enabling users to fully understand the operational mechanisms of the model.
    \item We seek to reduce the high GPU overhead associated with pre-training or fine-tuning in current graph plus LLMs paradigms by utilizing a new model optimization approach, known as Iterative Training through Prompt Optimization.
    \item We validate the effectiveness of this method from multiple perspectives on real-world datasets.
\end{itemize}

\section{Preliminaries}
In this section, we introduce the essential concepts, notations, and problem settings considered in this research. Our primary focus is on the node classification task over text-attributed graphs, which represents a fundamental downstream task within the field of graph learning. We begin by defining text-attributed graphs.

\textbf{Text-Attributed Graphs}. Text-attributed graphs (TAGs) can be formally described as $\mathcal{G} = (\mathcal{V}, \mathcal{A}, \{\mathcal{X}_n\}_{n \in \mathcal{V}})$, where $\mathcal{V}$ represents a set of $\mathcal{N}$ nodes, $\mathcal{A} \in \mathbb{R}^{\mathcal{N} \times \mathcal{N}}$ is the adjacency matrix, and $\mathcal{X}_n \in \mathcal{D}^{\mathcal{L}_n}$ denotes a sequential text associated with each node $v_n \in \mathcal{V}$. Here, $\mathcal{D}$ is the dictionary of words or tokens, and $\mathcal{L}_n$ is the length of the sequence. In this paper, we focus on the problem of node classification in TAGs. Specifically, given a subset of labeled nodes $\mathcal{L} \subseteq \mathcal{V}$, the task is to predict the labels of the remaining unlabeled nodes $\mathcal{U} = \mathcal{V} \setminus \mathcal{L}$.And iterates over the input mini-batch \(\mathcal{B}\) one-pass input.

\textbf{One-hop neighbors}. Given a node $v_i \in \mathcal{V}$, the set of one-hop neighbors, denoted as $\mathcal{N}(v)$, where $\mathcal{N}(v_i)=\{v_j \in \mathcal{V} | (v_i,v_j) \in \mathcal{E} \}$

\textbf{$k$-hop neighbors}. Given a node $v_i$, for $k \geq 2$, the $k$-hop neighbors of $v_i$ can be denoted as $\mathcal{N}^{k}(v_i)$, where $\mathcal{N}^k(v_i)=\left\{v_j \in \mathcal{V} \mid \exists v_m \in \mathcal{N}^{k-1}(v_i),(v_m, v_j) \in \mathcal{E} \land v_j \notin \mathcal{N}^{k-1}(v_i)\right\}$.

\section{Related Work}
In this section, we review the existing literature related to integrating Large Language Models (LLMs) and Graph Neural Networks (GNNs). Prior work has focused on several key areas, including traditional methods for trusted GNNs, the role of LLMs in graph-based tasks, and recent advances in optimization frameworks utilizing LLMs. We explore these approaches to highlight their contributions and limitations, establishing the foundation for our proposed Verbalized Graph Representation Learning (VGRL) framework. 

\subsection{Graph and LLMs}

\textbf{Traditional approaches to trusted GNNs.}
There are currently two main approaches: post-hoc explanation methods and self-interpretable models . The former tries to interpret the output of the model by adding a model-independent interpreter, for example ~\citep{ying2019gnnexplainer, vu2020pgm, zhang2023rsgnn}. However, this can lead to incomplete explanatory information in the output, or even generate explanatory information that is incorrect in the opinion of humans. The latter tries to solve this problem by constructing models that themselves have interpretable principles, for example ~\citep{dai2021towards, zhang2022protgnn}. However, these interpretable principles are based on their inductive bias, and only experts in the relevant fields can accurately judge whether such inductive bias is reasonable or not.

\textbf{LLM in Graph.}
Existing methods are mainly categorized into three types: (1) LLM as Enhancer which mainly enhances the performance of GNNs by adding LLM-generated information, for example ~\citep{he2023harnessing, chen2024exploring, ni2024enhancing}; (2) LLM as Predictor which mainly performs a downstream task by directly inputting the graph structure into the LLM, for example ~\citep{tang2024graphgpt, qin2023disentangled}; (3) LLM as Alignment which mainly enhances the performance by aligning embedding spaces of GNNs and LLMs, for example ~\citep{yang2021graphformers, mavromatis2023train}. Among them, there is explanation-based LLM-as-Enhancer approach ~\citep{he2023harnessing}, which achieves better performance by letting LLM generate natural language explanation information of graph structures and then embedding it into GNNs for downstream tasks. However, after the embedding from natural language to graph structure is not directly visible as a black box to humans, and can only be proven effective indirectly through the performance of downstream tasks.

\subsection{LLMs Optimization}
\textbf{LLMs for planning and optimization}. Large language models (LLMs) have been successfully applied to planning tasks for embodied agents ~\citep{song2023llm,xie2023translating,li2022pre,liang2023code}, enabling them to follow natural language instructions and complete complex tasks. More recently, LLMs have also been utilized to tackle optimization problems by generating new solutions from prompts that incorporate previously generated solutions and their associated loss values. While these LLM-based optimization ~\citep{xiao2024verbalized,yang2024zhongjing} methods bear some resemblance to our approach, as we also use LLMs to address optimization challenges, a key limitation of existing work is that it has not yet been explored in the graph domain. To address this gap, we propose an extension of this framework to the graph domain, introducing Verbalized Graph Representation Learning (VGRL), which applies LLMs to graph neural networks (GNNs) and opens new possibilities for solving graph-based optimization problems through natural language interactions.

\textbf{Prompt engineering and optimization}. Numerous prompting techniques ~\citep{wei2022chain,zhang2022automatic,zhou2022large,wang2022self,yao2024tree,yao2023beyond,weston2023system} have been developed to enhance the reasoning capabilities of LLMs. To minimize the manual effort required in designing effective prompts, various automatic prompt optimization approaches ~\citep{zhang2022automatic,zhou2022large,yang2024zhongjing,pryzant2023automatic,wen2024hard,deng2022rlprompt,li2024guiding,ma2024large,sordoni2023deep} have been introduced. However, traditional prompt optimization methods primarily focus on refining the text prompt without changing its underlying semantic meaning. In contrast, our VGRL framework goes beyond mere prompt adjustments by directly updating the parameters of the language-based model through the integration or modification of prior information. This not only improves optimization but also ensures that the learner model remains fully interpretable in its predictions, offering a more robust and transparent solution for graph-based learning tasks.

\textbf{LLMs for multi-agent systems}. Given their strong instruction-following capabilities, LLMs can assume various roles within multi-agent systems \citep{qian2023communicative,wu2023autogen,hong2023metagpt,li2023camel}. For instance,  explore multi-agent collaboration systems designed to solve complex tasks such as software development. In the VGRL framework, this concept is extended to a two-agent system, where one LLM functions as the learner and the other as the optimizer.

Our approach sidesteps the problem of modeling black boxes by having the LLM generate human-readable information as promt of another LLM making it perform the downstream task. This can be viewed as a “guidance-feedback-redirection” process between models, which, after many iterations, returns the optimal guidance solution for a given task, which is directly human-readable.

\begin{figure}
    \centering
    \includegraphics[width=\linewidth]{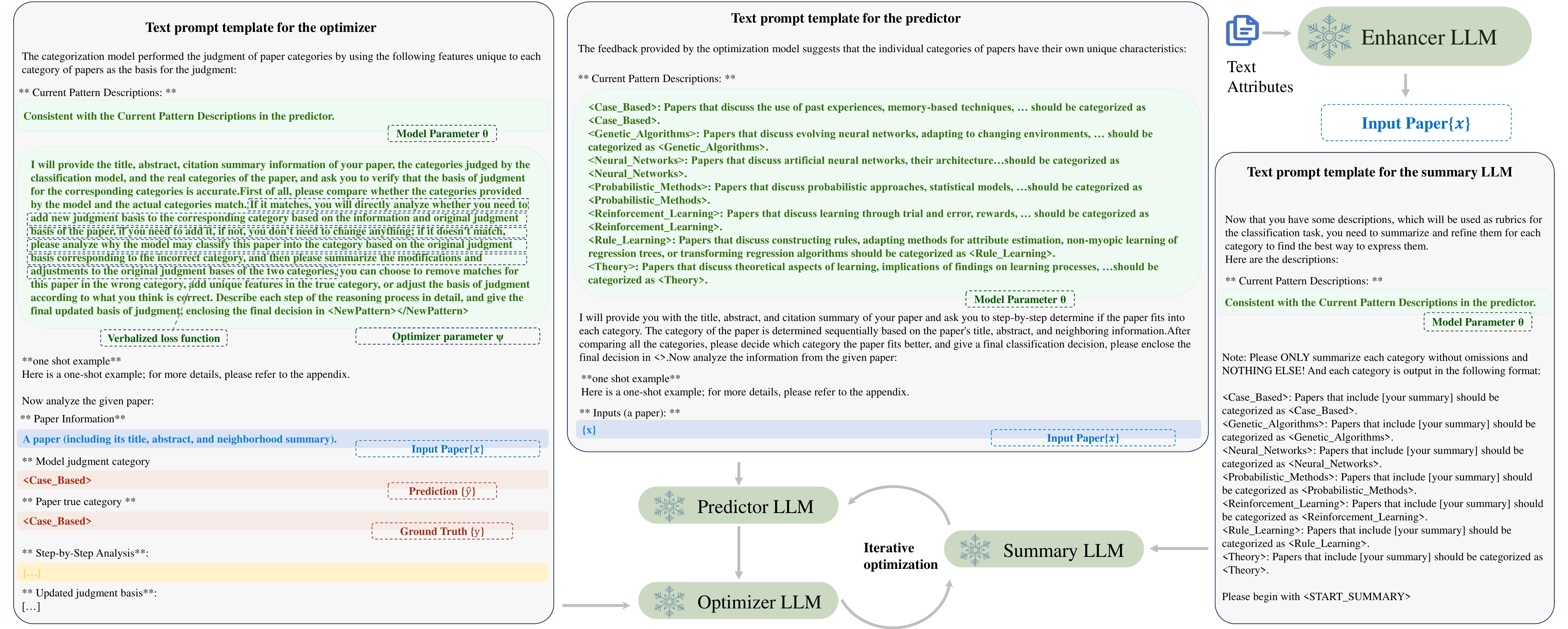}
    \caption{An overview of iterative optimization and text prompt templates for the predictor, optimizer, and summary LLM in the node classification example}
    \label{fig:framework}
\end{figure}

\section{Proposed Method}

In this paper, we present the Verbalized Graph Representation Learning (VGRL) framework, a pioneering approach that integrates large language models (LLMs) with graph-based tasks while ensuring full interpretability throughout the process. Our methodology encompasses four innovative components, each designed to enhance both the performance and the transparency of LLMs in handling graph data.

\subsection{Incorporating Graph Structure into LLM Predictions}
Although Large Language Models (LLMs) can achieve competitive zero-shot performance on specific datasets without considering graph structures, their performance often lags behind Graph Neural Networks (GNNs) on benchmark datasets such as CORA, CITESEER, and OGBN-ARXIV. This gap underscores the importance of graph structural information in tasks like node classification, indicating the need to explore how incorporating graph structures into prompts could enhance LLM performance.

Given that LLMs (e.g., ChatGPT) are not natively designed to process adjacency matrices or perform graph-based computations, it is impractical to directly integrate graph operations into LLMs. Thus, an alternative approach is to verbalize graph information as text that LLMs can process effectively. This transformation allows LLMs to interpret node relationships and structural dependencies in natural language format. In \citep{chen2024exploring}, various methods are evaluated to represent node connections textually, aiming to enhance LLM reasoning capabilities for graph-based tasks.

One effective method is the `ego-graph' approach, which focuses on the local subgraph surrounding a target node. By constraining the LLM's focus to a limited number of nodes, this method reduces complexity while preserving key local graph structure. To simulate the neighborhood aggregation process typical in GNNs, the input prompt incorporates a summary of attributes from neighboring nodes. Thus, important information from the graph is conveyed to the LLM without altering its reasoning mechanisms. This process can be formalized as:
\begin{equation}
    Z_{v_i}^1 = f_e\left(\mathcal{X}_{v_i}, \left\{\mathcal{X}_{v_j} \mid v_j \in \mathcal{N}(v_i)\right\}\right)
\end{equation}
where \( Z_{v_i}^1 \) is the enhanced representation of node \( v_i \) with one-hop neighbor information, \( \mathcal{X}_{v_i} \) represents the features of node \( v_i \), and \( \mathcal{N}(v_i) \) denotes the set of one-hop neighbors of \( v_i \). The function \( f_e \) encapsulates the process of verbalizing neighborhood information and processing it by the LLM.

Inspired by this ego-graph approach, we have also introduced a method for incorporating structural information into our model. By embedding the attributes and relationships of neighboring nodes into the prompt, we aim to enable the LLM to better capture the interactions between nodes. Below is an example of a neighbor summary in Table \ref{table:nsumm}:

\begin{table}[h!]
    \caption {Prompts used to generate neighbor summary.}
    \label{table:nsumm}
    \centering
    \rule{\textwidth}{2pt}
    \parbox{\textwidth}{
    \vspace{5pt}
    \textbf{Prompts used to summarize the neighboring information} \\
    I will now give you basic information about all the papers cited in a paper; this information includes: the abstracts and categories of the cited papers.
    The following list records some papers related to the current one. \\
    \text{[}\{ "content": "This paper firstly provides ...", "category": "Rule Learning"... \}, ...\text{]}   \\
    \# \textbf{Instruction} \\ 
    Please summarize the information above with a short paragraph, find some common points which can reflect the category of this paper.
    
    Note: ONLY your summary information and NOTHING ELSE!
    
    Please start with "The papers cited in this essay".
    \vspace{2pt}
    }
    \rule{\textwidth}{2pt}
    \vspace{5pt}
\end{table}

\subsection{Verbalizing Model Parameters for Interpretability}

Traditional machine learning models, such as neural networks, rely on numerical parameters, $\theta = \{\theta_1, \theta_2, \dots, \theta_t\}$, which are often difficult to interpret. These parameters are typically represented as abstract numerical values, making it complex and non-intuitive to understand or explain the internal workings of the model. In contrast, the Verbalized Graph Representation Learning (VGRL) framework leverages large language models (LLMs) to express model parameters through natural language, providing full interpretability.

In VGRL, the model parameters $\theta_t$ are defined by a text prompt, which consists of human-readable natural language tokens, $\theta_t \in \Theta_{\text{language}}$, where $\Theta_{\text{language}}$ is the set of all interpretable text sequences. This approach contrasts with traditional models where parameters are abstract numbers, which are hard to interpret directly. The VGRL framework unifies both data and model parameters into a natural language-based format that is inherently understandable.

The key features of this framework include:

\begin{itemize}
    \item \textbf{Discrete Parameters}: The natural language used to express parameters $\theta$ is inherently discrete. This is in contrast to the continuous parameter representations in traditional models, enhancing the intuitiveness of parameter interpretation.
    \item \textbf{Sequential Structure}: The parameters exhibit a sequential structure, as $\theta = \{\theta_1, \theta_2, \dots, \theta_t\}$, reflecting the temporal or contextual relationships between parameters. This sequential nature aids in capturing and understanding the dynamics between parameters.
    \item \textbf{Human Interpretability}: Since the parameters $\theta_t$ are verbalized in natural language, they are inherently comprehensible to humans. This allows the model's reasoning process and learning mechanisms to be more transparent, facilitating interpretability and easier analysis.
\end{itemize}

An advantage of using natural language for model parameters is that it enables the integration of prior knowledge and inductive biases directly into the model. As the model updates its parameters $\theta_t$, the changes are fully interpretable, providing clear insights into what the model is learning. For example, changes in $\theta_t$ can be directly mapped to natural language descriptions, offering an intuitive understanding of the model’s learning process.

Our empirical evidence demonstrates that text-based parameters often correspond to recognizable patterns in the data, further reinforcing the interpretability and transparency of the VGRL approach. This natural language parameterization not only enhances the intuitiveness of model but also improves its application, offering clearer insights into model tuning and interpretation in real-world scenarios.

\subsection{Leveraging LLMs for Node Classification}
Our approach centers on utilizing LLMs as interpretable predictors by querying them in an `open-ended' manner. Unlike existing methods that primarily rely on message passing mechanisms, our method employs a label feature matching mechanism. We match based on the inherent characteristics of the nodes themselves and the information from their neighbors. This label feature matching mechanism places a stronger emphasis on the intrinsic attributes of node, as it aligns with the insights provided in the prompt. 

The core of this method is represented by the following equation:

\begin{equation}
    \hat{y}_{v_i} = f_{p}(Z_{v_i}^k, \theta_{t-1})
\end{equation}

Here, \( \hat{y}_{v_i} \) denotes the predicted label for node \( v_i \), and \( Z_{v_i}^k \) represents the enhanced node representation incorporating \( v_i \)'s \( k \)-hop neighbors. \( \theta_{t-1} \) refers to the LLM's parameters at the previous step, enabling the model to leverage its prior knowledge and reasoning capabilities to generate the prediction. The function \( f_{p} \) serves as the predictor that utilizes the enhanced representation and model parameters to produce the label output. This formulation emphasizes the LLM’s role as a predictor, focusing on generating interpretable outputs.

For each node \( v_i \in \mathcal{V} \), a prompt is crafted that includes not only the node's features, such as the paper title and abstract, but also relevant graph structure information. Specifically, the attributes of neighboring nodes up to the \( k \)-hop neighborhood are embedded in the prompt, as encapsulated in \( Z_{v_i}^k \). This enables the LLM to better understand the node's context and surroundings within the graph, leading to more informed and accurate predictions.

\subsection{LLM as an optimizer with interpretable optimization process}
For the predictor LLM, we provide textual descriptions of node categories, which serve as model parameter $\theta$, and the model determines which category the input node $v_i$ belongs to based on the given descriptions. The quality of node category descriptions $\theta$ directly affects the performance of LLM predictions; hence, obtaining suitable node category descriptions is very important. Additionally, for better explainability, VGRL imposes a strong constraint on $\theta$, ensuring that the updated $\theta$ still belong to natural language sequences that humans can understand. 

Under these conditions, it is not advisable to use classical machine learning optimization methods such as gradient descent to optimize $\theta$. Inspired by \cite{xiao2024verbalized}, the optimizer LLM can output natural language that satisfies the constraints, so we only need to ask the LLM to play the role of an optimizer, then optimized category descriptions are also in natural language understandable by humans. Therefore, we directly use another LLM to optimize $\theta$. Given a mini-batch $\mathcal{B}$, the optimization process is as follows:
\begin{equation}
    \widetilde{\theta}_{v_i}^t = {g_{opt}}(Z_{v_i}^k, y_{v_i}, \hat{y}_{v_i}, \theta_{t-1}, \Psi), v_i \in \mathcal{B}
\end{equation}
where $y_{v_i}$ is the true label of $v_i$, $\widetilde{\theta}_{v_i}^t$ represents the intermediate parameter values for node $v_i$ during the $t$-th iteration, and $\Psi$ denotes the parameter of the optimizer LLM, which is a text prompt. Specifically, we optimize the intermediate parameter value $\widetilde{\theta}_{v_i}^t$ of each node $v_i$ in $\mathcal{B}$, and then summarize the intermediate parameter values of these nodes through a summary LLM (Section \ref{section:summary}) to obtain a new round of parameter $\theta_t$.
The overall framework for optimizer optimization and the text prompt template are given in Figure \ref{fig:framework}.  The parameter $\Psi$ of the optimizer LLM is actually a text prompt provided by humans and is not updated. The text prompt linguistically specifies the optimization loss function, guiding the optimizer LLM to optimize $\theta$. The LLM-parameterized optimizer allows users to interact with it directly, which not only helps to trace model failures but also permits the incorporation of prior knowledge to enhance optimization.
In addition, we also guide the LLM to output explanations of the optimization process, demonstrating the explainability of the VGRL optimization process.

\subsection{Summary LLM}
\label{section:summary}
The role of the Summary LLM is to aggregate and summarize the updated intermediate parameters from the optimizer LLM, generated during the previous minibatch, to obtain updated \( \theta \). Specifically, given a set of updated parameters from the last minibatch \( \mathcal{B} \), the Summary LLM consolidates these updates into a new set of parameters, \( \theta_t \). This process can be formalized as:

\begin{equation}
    \theta_t = f_{s}\left(\{\widetilde{\theta}_{v_i}^t \mid v_i \in \mathcal{B}\}\right)
\end{equation}

Here, \( \widetilde{\theta}_{v_i}^t \) represents the intermediate parameter values for node \( v_i \) during the \( t \)-th iteration, and \( \mathcal{B} \) denotes the set of nodes in the current minibatch. The function \( f_{s} \) operates by combining these parameter updates to produce a cohesive set of parameters, \( \theta_t \), which reflects the overall learning progress across the minibatch. This aggregation ensures that key information from each node's updated parameters is captured while maintaining coherence in the overall optimization process.

\subsection{Chain-of-Thought Prompting}
Inspired by ~\citep{wei2022chain}, we introduce the zero-shot and one-shot Chain-of-Thought (CoT) methods in prompt. For the zero-shot method, we encourage the LLM to perform step-by-step text generation by restricting and guiding the LLM to make the generated explanatory information as structured and precise as possible, in order to achieve a better final result generation based on the self-generated information.  Although zero-shot VGRL is already fully interpretable, we still want to customize the interpretation in specific domains to ensure that the interpretation information is more in line with the norms of the human mind and thus enhance the model's performance. Therefore, we introduce the one-shot method by manually constructing a sample of the CoT, so that the model can generate the interpretation information and the final output based on the sample. The motivation for the one-shot approach is that we believe that the content generated by the LLM based on a sample that conforms to the logic of the human mind will better contribute to the completion of the final task.

\section{Experiments}
In this section, We will compare the performance of the VGRL framework with diverse backbone models for the TAG node classification task.We will answer the following questions:

\begin{itemize}
    \item \textbf{Q1:} Can VGRL framework increase the performance of backbone models?
    \item \textbf{Q2:} Do each part of the VGRL framework play a positive role?
\end{itemize}

\subsection{Baseline and Experiment Setting}
We use two LLM-as-predictor models as backbones ~\citep{chen2024exploring}, and add our framework on top of them for comparisons. Information on our equipment can be found at Table \ref{table:devices}.

\begin{itemize}
    \item \textbf{Node only:} `node only' refers to the features considering only the node itself, excluding any neighbor information.
    \item \textbf{Summary}: `Summary' indicates that we used an independent LLM to summarize the node’s $k$-hop information, which can be viewed as the introduction of an enhancer LLM for encoding the graph structure. The prompt for the enhancer LLM is shown in Table \ref{table:nsumm}.
\end{itemize}

\begin{wraptable}{r}{0.45\textwidth}
    \centering
    \footnotesize
    \caption{Information on our equipment}
    \resizebox{0.42\textwidth}{!}{
        \begin{tabular}{c|c}
        \hline
        \multicolumn{2}{c}{Devices} \\
        \hline
        OS         & Ubuntu 22.04.4 LTS x86\_64             \\
        Language   & Python 3.10.14                         \\
        Frameworks & pytorch 2.4.0 + cuda 12.4              \\
        CPU        & Intel Xeon Silver 4310 (48) @ 3.300GHz \\
        GPU        & 3 * NVIDIA L20 (48G)                       \\
        Memory     & 128508MiB                              \\
        \hline
        \end{tabular}
    }
    \label{table:devices}
\end{wraptable}

During the experiments, we used one-hop neighbor information for summarization and set model temperature $\tau=0.1$ as default. Additionally, we introduced prior knowledge in our comparison by manually constructing prior knowledge as the initial optimize $\theta$ for iterative processing. And we setting a mini-batch training process with a batch size of 8, i.e. $|\mathcal{B}| = 8$.

\subsection{Main Results (Q1)}
We conducted evaluations on the Cora TAG ~\citep{McCallumIRJ} dataset (See ~Appendix\ref{dataset_desc}) by comparing our optimization iterative process with the baseline that excludes the VGRL framework ~\citep{chen2024exploring}. The results are presented in Table \ref{table:main}. We extracted a subset of nodes from the Cora dataset as our experimental data. For further steps, we blurred the concept of epochs and treated each batch as a single step.

\begin{minipage}{0.6\textwidth}
    \centering
    \captionof{table}{Node classification accuracy for the Cora dataset}
    \tiny
    \resizebox{1\textwidth}{!}{
    \begin{tabular}{c|c c|c c}
        \hline
           \multirow{2}{*}{Cora} & \multicolumn{2}{c|}{w/ prior} & \multicolumn{2}{c}{w/o prior} \\
        \cline{2-5} 
         & zero-shot & one-shot & zero-shot & one-shot \\
        \hline
        Node only & 0.625 & 0.400 & \textbf{0.675} & 0.100 \\
        Node only + VGRL & \textbf{0.650} & \textbf{0.625} & \textbf{0.675} & \textbf{0.475} \\
        \hline
        Summary & 0.650 & 0.550 & 0.700 & 0.475 \\
        Summary + VGRL & \textbf{0.800} & \textbf{0.700} & \textbf{0.875} & \textbf{0.700} \\
        \hline
    \end{tabular}
    }
    \label{table:main}
    \centering
    \captionof{table}{Ablation study on the Cora dataset, showing the effects of different variants base on Summary + VGRL on the accuracy performance}
    \tiny
    \resizebox{1\textwidth}{!}{
    \begin{tabular}{c|c c|c c}
        \hline
           \multirow{2}{*}{Cora Summary + VGRL} & \multicolumn{2}{c|}{w/ prior} & \multicolumn{2}{c}{w/o prior} \\
        \cline{2-5} 
         & zero-shot & one-shot & zero-shot & one-shot \\
        \hline
        original method & \textbf{0.800} & \textbf{0.700} & \textbf{0.875} & \textbf{0.700} \\
        w/o optimizer LLM & 0.650 & 0.550 & 0.700 & 0.475 \\
        w/o summary LLM & 0.650 & 0.625 & 0.725 & 0.625 \\
        \hline
    \end{tabular}
    \label{table:ablation}
    }
\end{minipage}%
\begin{minipage}{0.4\textwidth}
    \centering
    \begin{tikzpicture}[scale=0.65]
    
        \begin{axis}[
            xlabel=Step,
            ylabel=Test Accuracy,
            tick align=outside,
            legend columns=2,
            legend style={at={(0.5,-0.2)},anchor=north}
        ]
            
            \addplot[smooth,mark=*,color={rgb,255: red,102; green,204; blue,255}] plot coordinates {
                (0, 0.65) (5, 0.775) (10, 0.8) (15, 0.8)
                (20, 0.675) (25, 0.75) (30, 0.75) (35, 0.725)
                (40, 0.65) (45, 0.775) (50, 0.75) (55, 0.775)
                (60, 0.6) (65, 0.5) (70, 0.775) (75, 0.775)
                (80, 0.8)
            };
            \addlegendentry{zero-shot-with-prior}
            
            \addplot[smooth,mark=triangle,color={rgb,255: red,216; green,0; blue,0}] plot coordinates {
            (0, 0.7) (5, 0.75) (10, 0.8) (15, 0.8)
            (20, 0.8) (25, 0.875) (30, 0.75) (35, 0.75)
            (40, 0.8) (45, 0.725) (50, 0.75) (55, 0.775)
            (60, 0.75) (65, 0.75) (70, 0.8) (75, 0.8)
            (80, 0.75)
            };
            \addlegendentry{zero-shot-wo-prior}
            
            \addplot[smooth,mark=*,color={rgb,255: red,255; green,192; blue,203}] plot coordinates {
            (0, 0.55) (5, 0.6) (10, 0.675) (15, 0.575)
            (20, 0.625) (25, 0.675) (30, 0.6) (35, 0.625)
            (40, 0.65) (45, 0.65) (50, 0.625) (55, 0.7)
            (60, 0.7) (65, 0.625) (70, 0.65) (75, 0.65)
            (80, 0.65)
            };
            \addlegendentry{one-shot-with-prior}
            
            \addplot[smooth,mark=triangle,color={rgb,255: red,57; green,197; blue,187}] plot coordinates {
            (0, 0.475) (5, 0.575) (10, 0.65) (15, 0.6)
            (20, 0.6) (25, 0.6) (30, 0.6) (35, 0.7)
            (40, 0.65) (45, 0.7) (50, 0.625) (55, 0.725)
            (60, 0.6) (65, 0.65) (70, 0.625) (75, 0.6)
            (80, 0.675)
            };
            \addlegendentry{one-shot-wo-prior}
        
        \end{axis}
    \end{tikzpicture}
    \captionof{figure}{Summary+VGRL Acc-Step}
    \label{fig:acc-step}
\end{minipage}

Our comparison reveals that our framework, through the iterative process, achieves better performance, demonstrating the effectiveness of the VGRL framework in representation learning. VGRL gradually refines the label features through repeated iterations, as shown in Figure \ref{fig:acc-step}, which illustrates the change in test accuracy during the mini-batch iterations. Additionally, we used the open-source Llama3.1 8B model for all experiments, which not only significantly reduced costs but also proved the optimization capability of the framework itself.

\subsection{Ablation Experiments (Q2)}
We conducted ablation experiments on the Summary + VGRL architecture to assess the importance and relevance of each module. The results of the ablation experiments are shown in Table \ref{table:ablation}.
\begin{itemize}
    \item \textbf{w/o optimizer LLM:} This variant removes the optimizer LLM, i.e., there is no iterative optimization process, which is equivalent to using the predictor LLM to make the final decision.
    \item \textbf{w/o summary LLM:} This variant removes the summary LLM, i.e., after each optimization update, instead of summarizing the information through the summary LLM, the results of a batch update are directly used in the next iteration.
\end{itemize}

\subsection{Case Study}
\begin{figure}[h]
    \centering
    \includegraphics[width=1\linewidth]{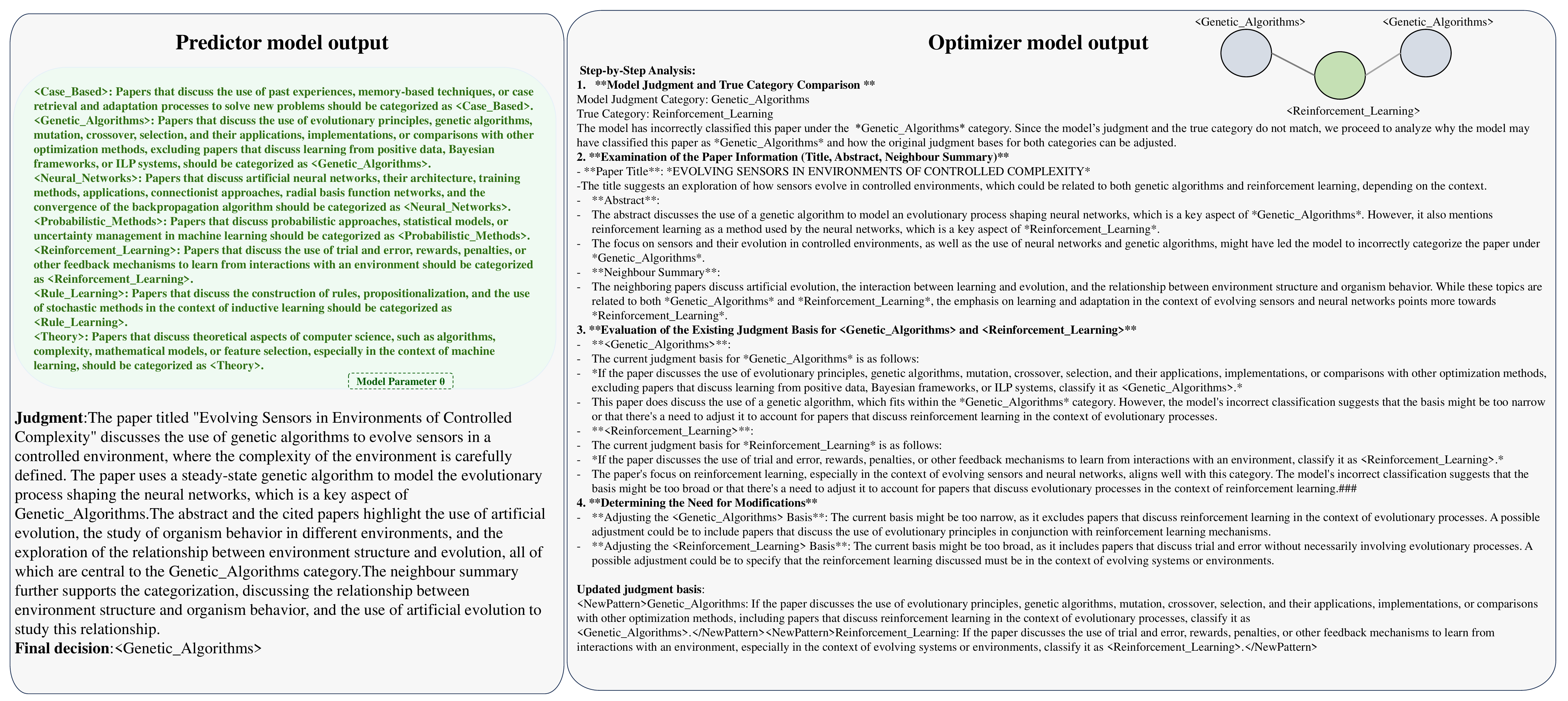}
    \caption{Case study for one-shot wo prior Summary + VGRL: (1) The left figure shows the explanation information and prediction labels output by predictor LLM; (2) The right figure shows the optimization process of optimizer LLM for the predicted content of predictor LLM in the left figure.(3) The top-right figure shows an example of the one-hop neighbors of a predicted sample.
}
    \label{fig:case_stduy}
\end{figure}

To explore the impact of the VGRL framework on the TAG node classification task, we conducted an analysis of a particular training sample from the Cora dataset, as shown in Figure \ref{fig:case_stduy}. In the paper `Evolving Sensors in Environments of Controlled Complexity' the one-hop neighboring nodes all have the label `Genetic\_Algorithms' while the actual label of the node is `Reinforcement\_Learning' This heterogeneity can significantly disrupt the node's feature information during neighborhood aggregation,  resulting in biased classification results. However, VGRL is able to effectively capture unique characteristics of each category, using them as a basis for matching the node's own features. This addresses the issue of information corruption caused by the propagation mechanism in heterogeneous graphs. 

Moreover, in the Cora dataset, paper categories cannot be strictly divided into binary classes. It is not uncommon for some nodes to belong to two categories simultaneously. In such cases, the label-feature matching mechanism proves to be more reasonable than the message-passing mechanism, as it focuses more on the node's own information (as can be inferred from the formulation of $\theta$). Making judgments and decisions based on one's existing knowledge ($\theta$) is the most fundamental decision-making process for humans.

`Judgment' and `Step-by-Step Analysis' represent the model's label matching process, which is also human-readable and interpretable. Whether its the Predictor LLM's process of analyzing the node's own features and supplementing it with neighborhood information, or the Optimizer LLM's analysis and adjustment of the two categories involved in classification errors, both demonstrate a complete and interpretable optimization process. The model explains each update iteration in detail, presenting it in human-readable language. With the help of the Summary LLM, the Predictor LLM and Optimizer LLM communicate and feedback effectively, ultimately constructing the best decision-making basis from scratch for the node classification task on the current dataset.

For a detailed training process see ~Appendix \ref{C} to ~Appendix \ref{G}.

\section{theoretical analysis}
In this section, our goal is to demonstrate that the category descriptions generated by LLM can provide useful information for predicting label categories. Specifically, if the obtained category descriptions can faithfully represent the information of each category, then they are useful. At the same time, the LLM is non-redundant, as it can provide information that $X$ cannot provide. Let $\theta$ be the textual category descriptions generated by LLM; $H_l$ are the embeddings of category from the LLM; $X$ are the input of graph structure embeddings, $y$ is the target and $H(\cdot|\cdot)$ is the conditional entropy. The specific proof process can be found in Appendix \ref{theorem}.
\begin{theorem}
    Given the following conditions:  
    1) Fidelity: $\theta$ can faithfully represent the information of $H_l$ such that $H(H_l | \theta) = \epsilon,$ with $\epsilon > 0$; 
    2)Non-redundancy: $H_l$ contains information not present in $X$, that is, 
    $H(y|X, H_l) = H(y|X) - \epsilon^{'}$, with $\epsilon^{'} > \epsilon$. Then it follows that $H(y|X, \theta) < H(y|X)$.
\end{theorem}

\section{Conclusion}
This paper introduces Verbalized Graph Representation Learning (VGRL), a novel approach to text-attributed graph learning that ensures full interpretability by representing learned parameters as textual descriptions instead of continuous vectors. This method enhances transparency and user understanding of the decision-making process, fostering greater trust in the model's outputs. While the current application is limited to foundational graph learning paradigms, VGRL shows promise for broader use in more complex models, offering potential advancements in explainable AI and graph-based learning systems.

\bibliography{refs}
\bibliographystyle{iclr2025_conference}
\clearpage
\appendix 

\section*{Appendix}

\section{Theoretical analysis}
\label{theorem}
In this section, our goal is to demonstrate that the category descriptions generated by LLM can provide useful information for predicting label categories. We formulate our theorem as follows:\\
\begin{theorem}
Given the following conditions:\\
    1) Fidelity: $\theta$ can faithfully represent the information of $H_l$ such that 
 \begin{equation}
 \label{eq:fidelity}
     H(H_l | \theta) = \epsilon,  \epsilon > 0;
 \end{equation}
    2)Non-redundancy: $H_l$ contains information not present in $X$, that is
    \begin{equation}
    \label{eq:non}
        H(y|X, H_l) = H(y|X) - \epsilon^{'},  \epsilon^{'} > \epsilon;
    \end{equation}
    Then we can obtain:
    \begin{equation}
    \label{eq:total}
        H(y|X, \theta) < H(y|X).
    \end{equation}
\end{theorem}
where $\theta$ be the textual category descriptions generated by $LLM$; $H_l$ are the embeddings of category from the $LLM$; $X$ are the input of graph structure embeddings, $y$ is the target and $H(\cdot|\cdot)$ is the conditional entropy.
    
\begin{proof}
We aim to demonstrate that $H(y|X, \theta) < H(y|X)$, the process is following:\\
Start with:
\begin{equation}
\label{eq:1}
    H(y|X, \theta)
\end{equation}
We decompose the original expression Equation \ref{eq:1} into two parts based on the properties of entropy:
\begin{equation}
\label{eq:2}
    H(y|X, \theta) = H(y|X,H_l,\theta) + I(y;H_l|X,\theta)
\end{equation}
Based on the definition of mutual information, we can obtain:
\begin{equation}
    \label{eq:3}
    I(y;H_l|X,\theta) = H(H_l|X, \theta)-H(H_l|X,\theta,y)
\end{equation}
Due to the non-negativity of conditional entropy, we have:
\begin{equation}
    \label{eq:4}
    I(y;H_l|X,\theta) \leq H(H_l|X, \theta)
\end{equation}
By substituting Equation \ref{eq:4} into Equation \ref{eq:2}, we further obtain:
\begin{equation}
    \label{eq:5}
     H(y|X, \theta) \leq H(y|X,H_l,\theta) + H(H_l|X, \theta)
\end{equation}
When conditional variables decrease, the conditional entropy increases; so we have: 
\begin{equation}
    \label{eq:6}
     H(y|X, \theta) \leq  H(y|X,H_l) + H(H_l|\theta)
\end{equation}
Applying the two aforementioned conditions and substituting Equations \ref{eq:fidelity} and \ref{eq:non} into Equation \ref{eq:5}, we can obtain:
\begin{equation}
     H(y|X, \theta) \leq H(y|X)+\epsilon-\epsilon^{'} < H(y|X)
\end{equation}
The conclusion is thus proven.
\end{proof}

\section{Dataset Description}\label{dataset_desc}
Cora ~\citep{McCallumIRJ}: The Cora dataset consists of Machine Learning papers. These papers are classified into one of the following seven classes: Case\_Based, Genetic\_Algorithms, Neural\_Networks, Probabilistic\_Methods, Reinforcement\_Learning, Rule\_Learning, Theory. The papers were selected in a way such that in the final corpus every paper cites or is cited by atleast one other paper. There are 2708 papers and 5429 links in the whole corpus.

\newpage
\section{one-shot CoT}\label{C}
The one-shot example.
\begin{figure}[h]
    \centering
    \tiny
    \includegraphics[scale=0.85]{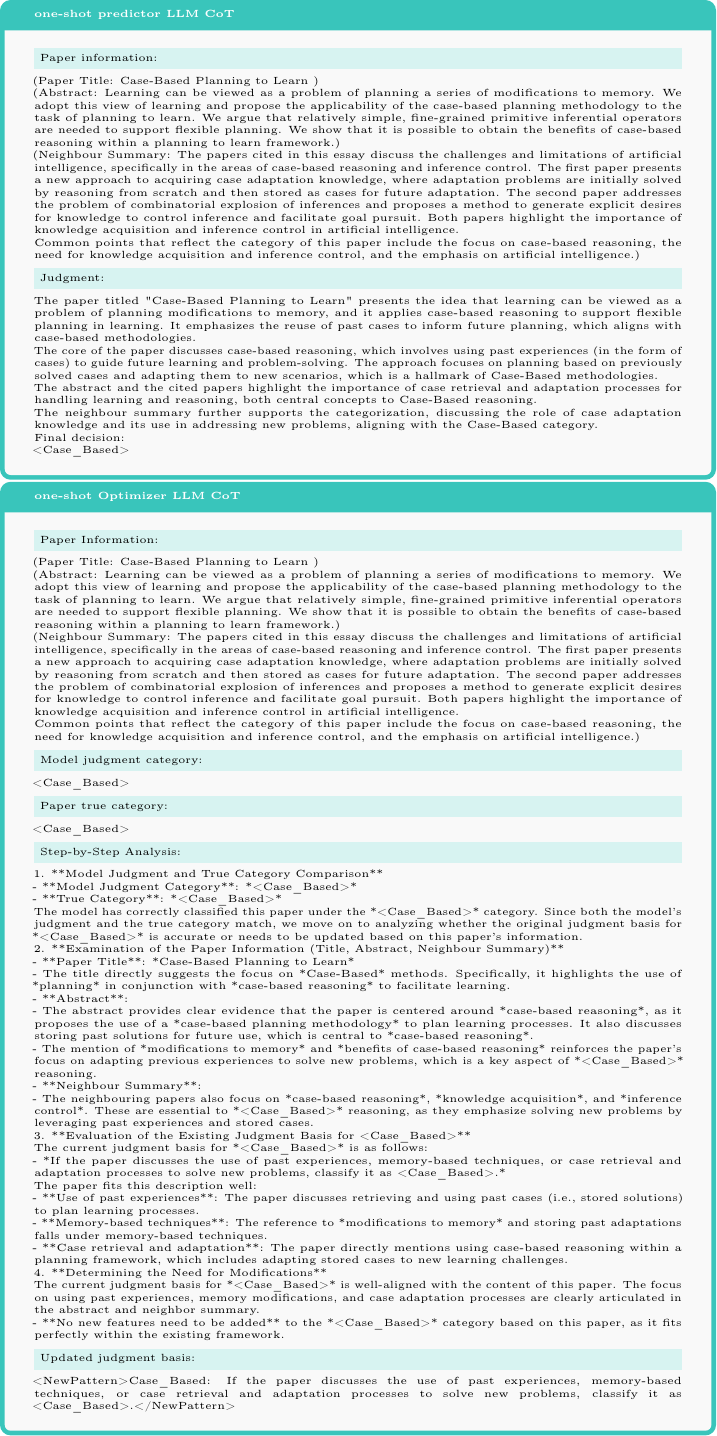}
    \label{fig:one-shot-CoT}
\end{figure}

\newpage
\section{Training Detail for Summary + VGRL}\label{D}
\subsection{one-shot w/o prior Summary + VGRL Step 1}
\begin{figure}[h]
    \begin{minipage}{0.4\textwidth}
        \centering
        \includegraphics[scale=0.5]{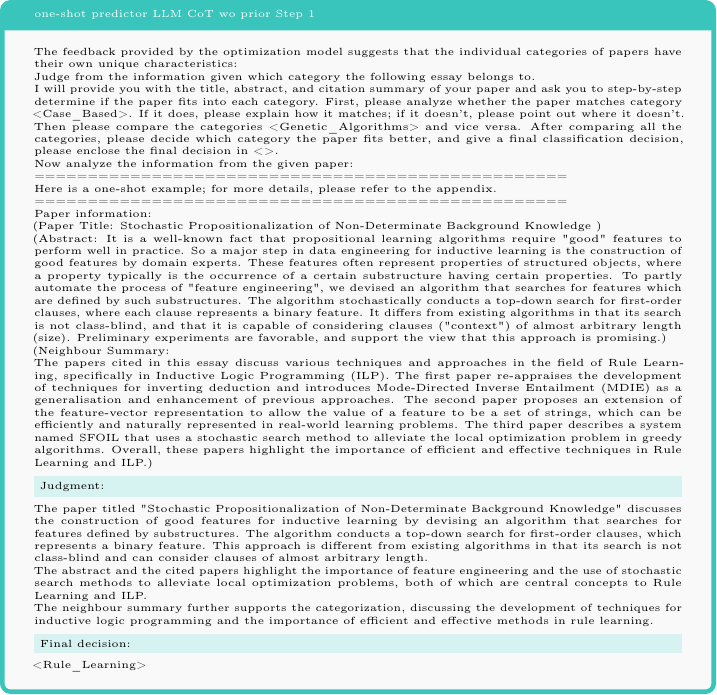}
        \label{fig:one-shot-pred-wop-1}
        \centering
        \includegraphics[scale=0.5]{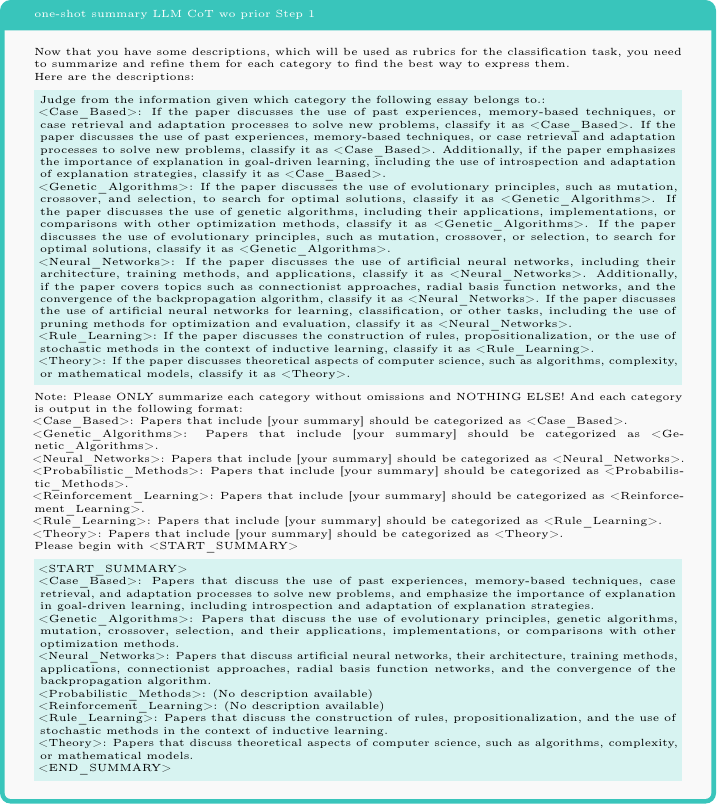}
        \label{fig:one-shot-pred-wop-1}
    \end{minipage}
    \hspace{0.5cm}
    \begin{minipage}{0.6\textwidth}
        \centering
        \includegraphics[scale=0.65]{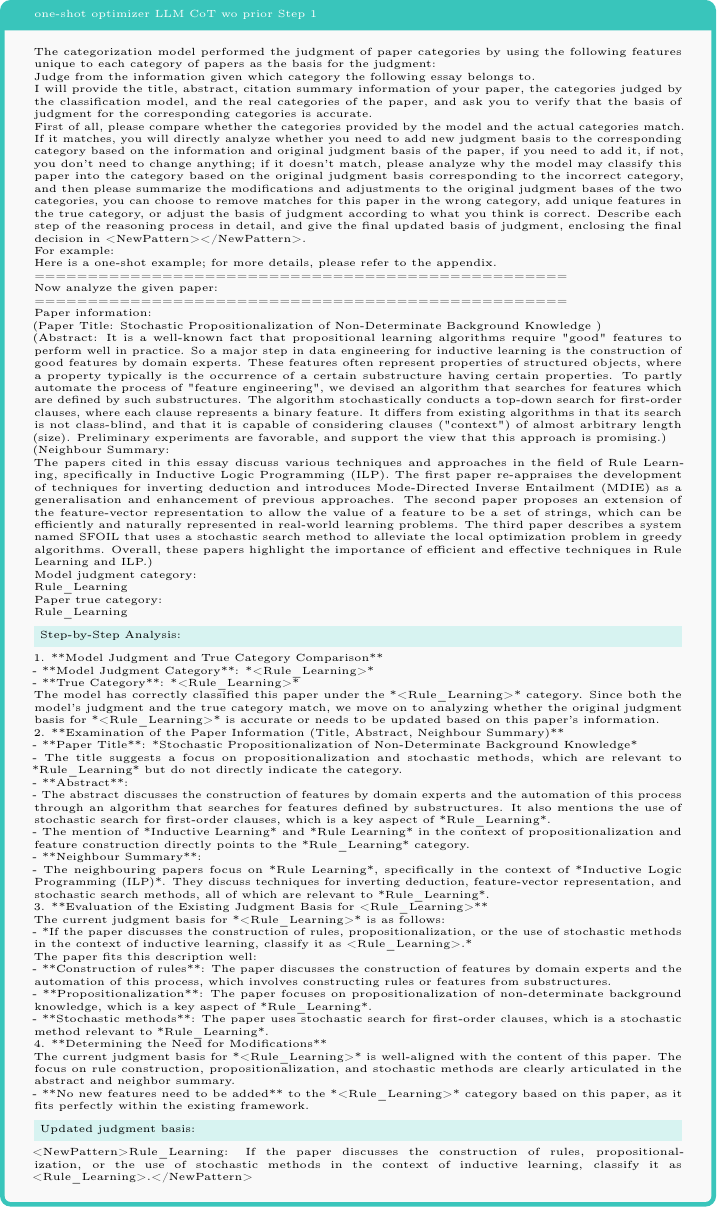}
        \label{fig:one-shot-pred-wop-1}
    \end{minipage}
\end{figure}

\newpage
\subsection{one-shot w/o prior Summary + VGRL Step 2}
\begin{figure}[h]
    \begin{minipage}{0.4\textwidth}
        \centering
        \includegraphics[scale=0.5]{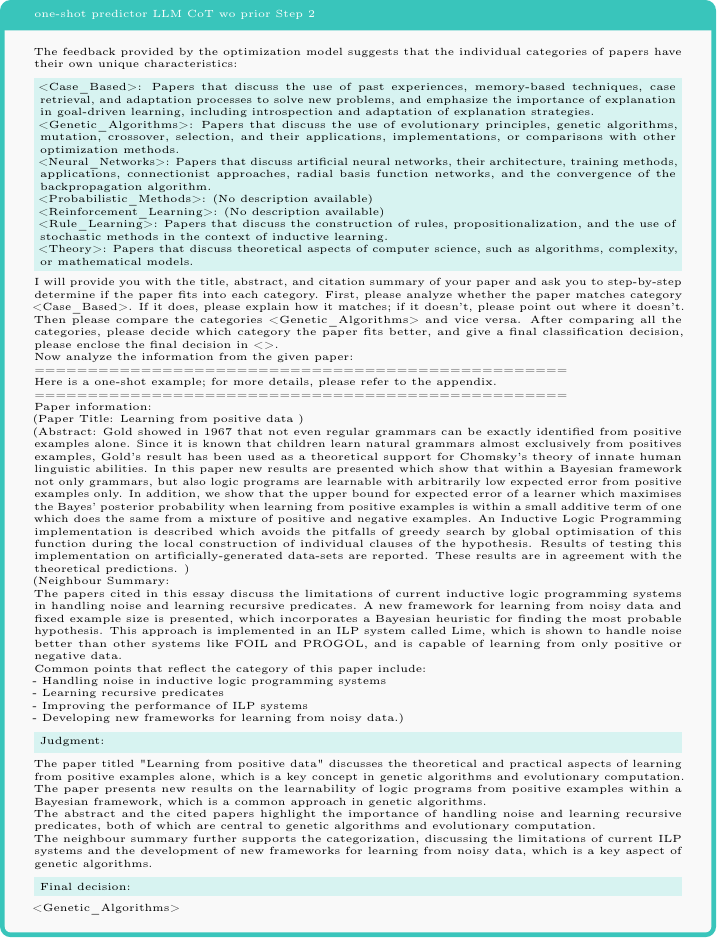}
        \label{fig:one-shot-pred-wop-2}
        \centering
        \includegraphics[scale=0.5]{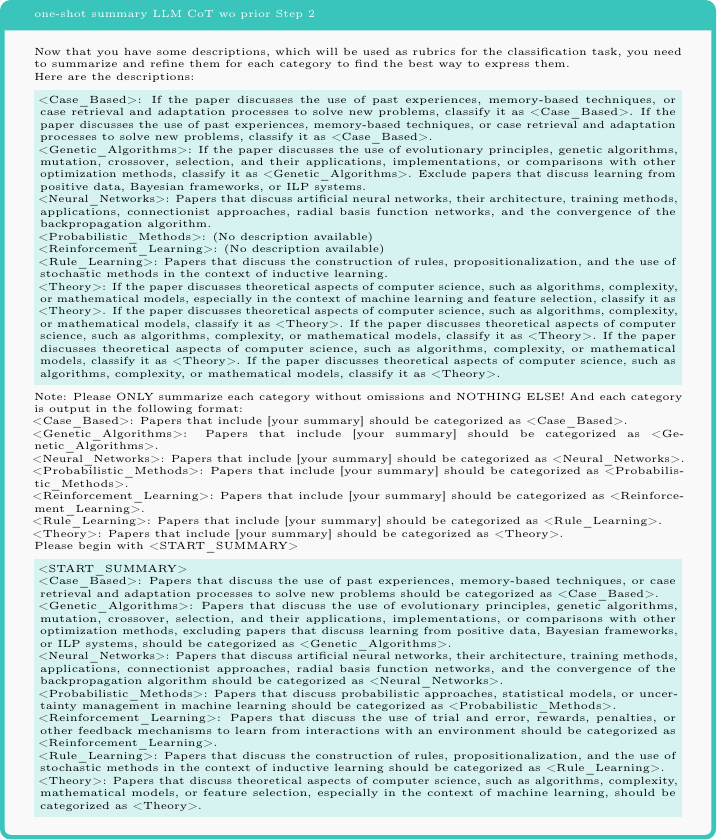}
        \label{fig:one-shot-pred-wop-2}
    \end{minipage}
    \hspace{0.5cm}
    \begin{minipage}{0.6\textwidth}
        \centering
        \includegraphics[scale=0.65]{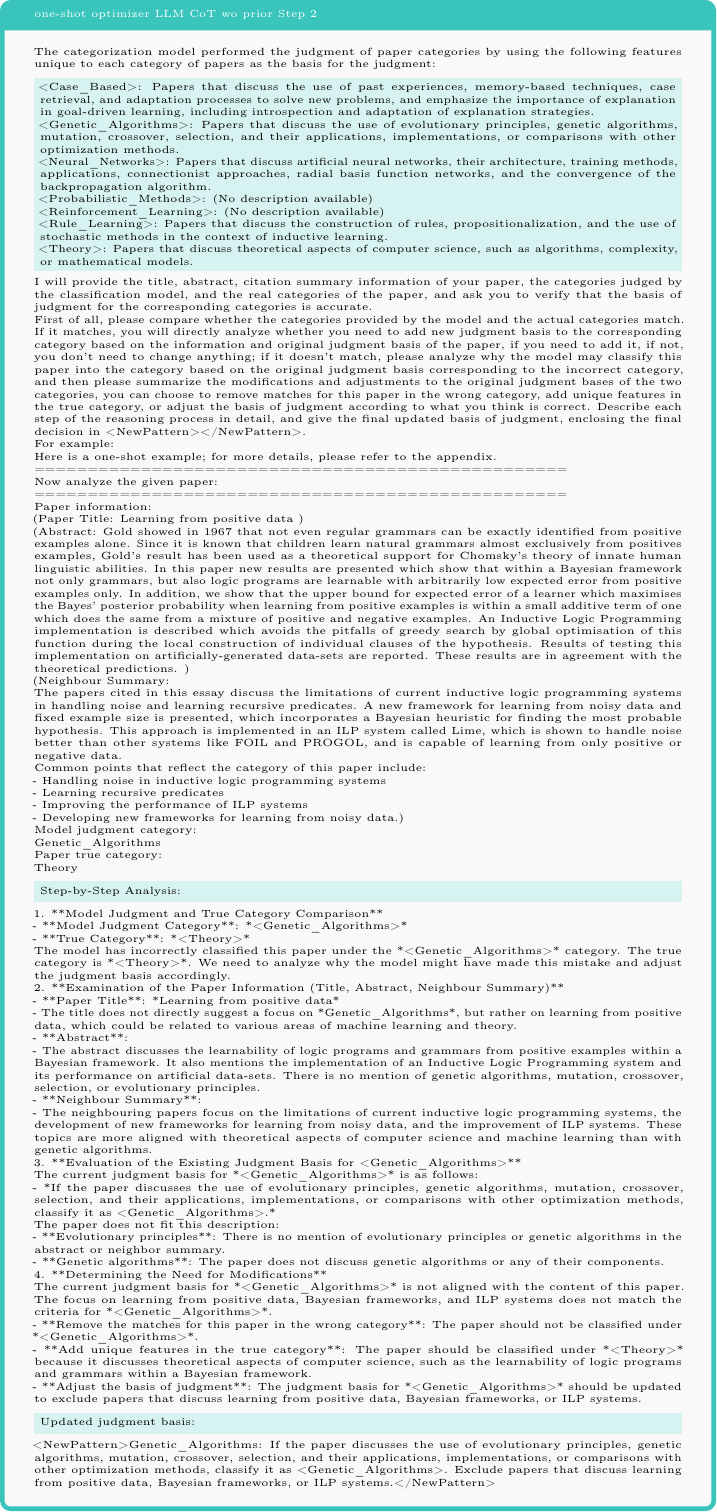}
        \label{fig:one-shot-pred-wop-2}
    \end{minipage}
\end{figure}

\newpage
\subsection{one-shot w/o prior Summary + VGRL Step 80}
\begin{figure}[h]
    \begin{minipage}{0.4\textwidth}
        \centering
        \includegraphics[scale=0.5]{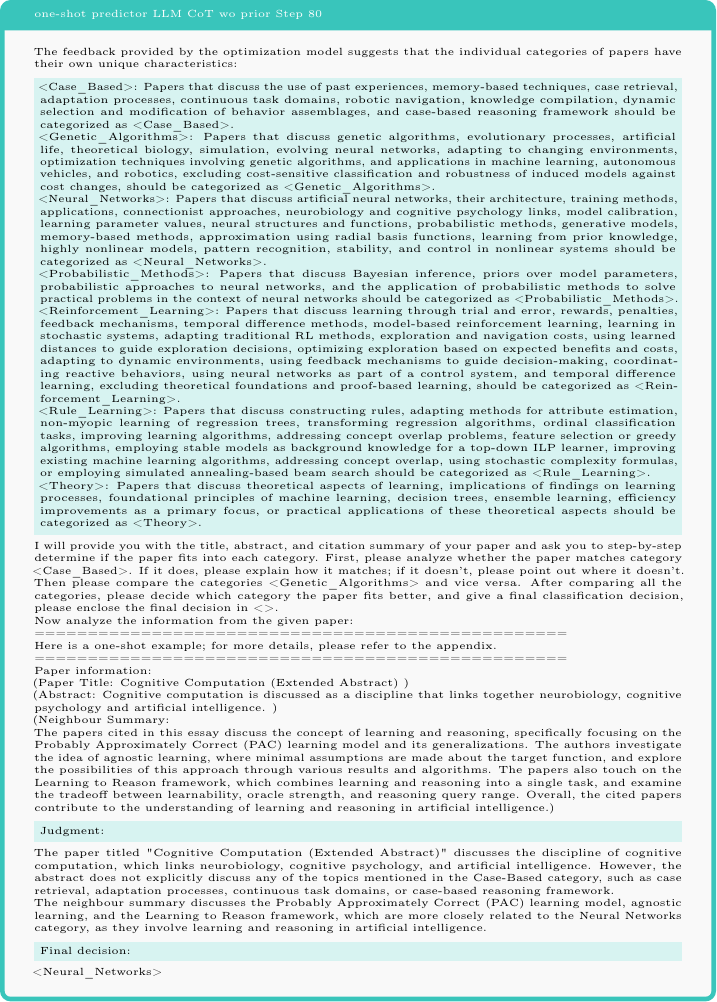}
        \label{fig:one-shot-pred-wop-80}
        \centering
        \includegraphics[scale=0.5]{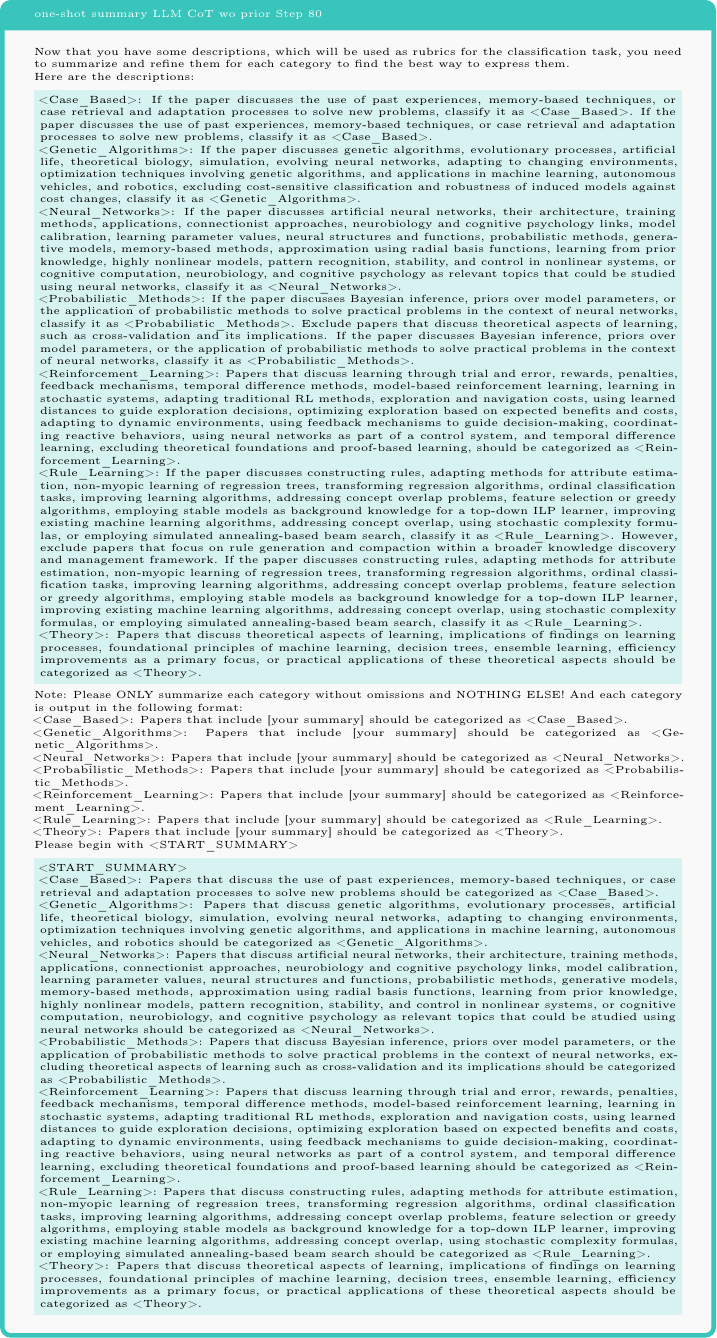}
        \label{fig:one-shot-pred-wop-80}
    \end{minipage}
    \hspace{0.5cm}
    \begin{minipage}{0.6\textwidth}
        \centering
        \includegraphics[scale=0.65]{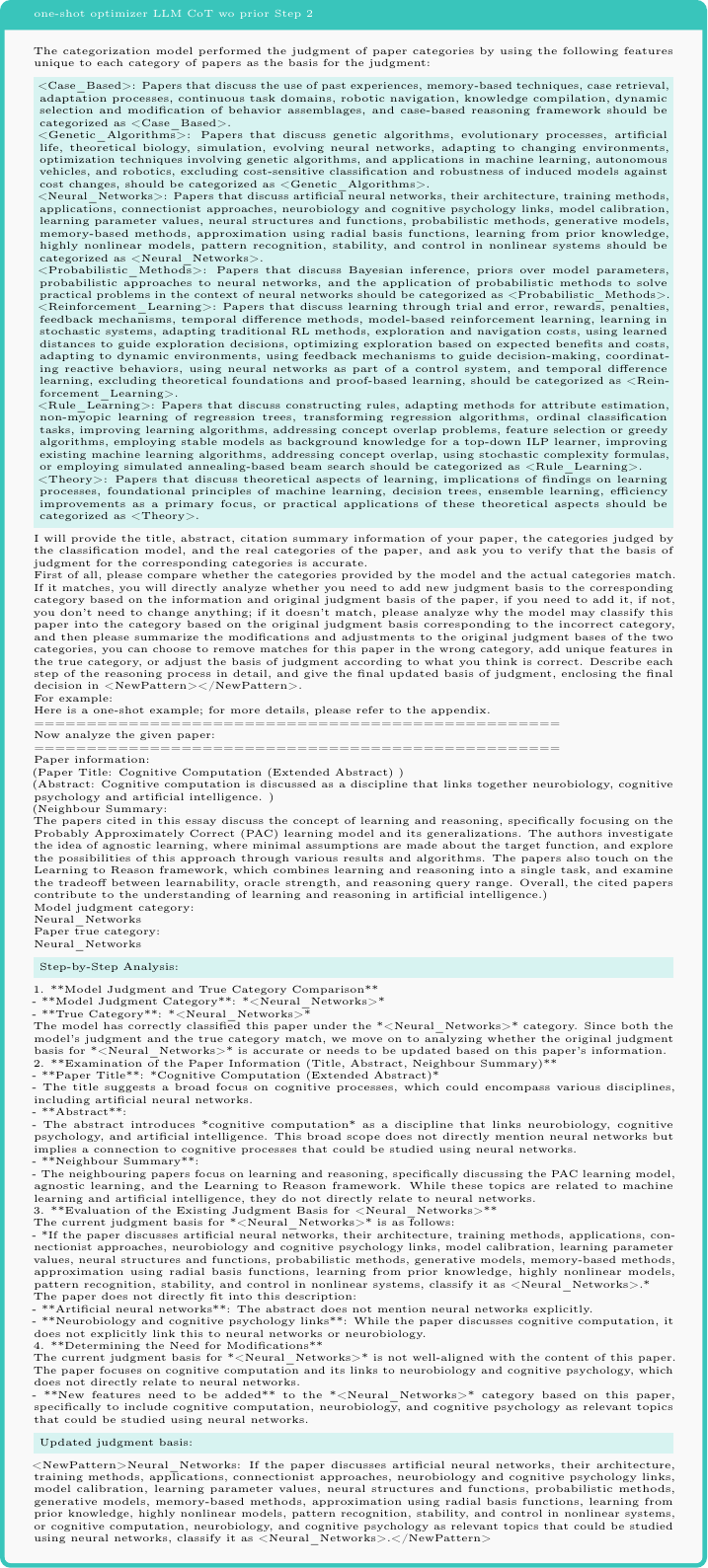}
        \label{fig:one-shot-pred-wop-80}
    \end{minipage}
\end{figure}

\newpage
\section{one-shot w/ prior Summary + VGRL}\label{E}
\subsection{one-shot w/ prior Summary + VGRL Step 1}
\begin{figure}[h]
    \begin{minipage}{0.4\textwidth}
        \centering
        \includegraphics[scale=0.5]{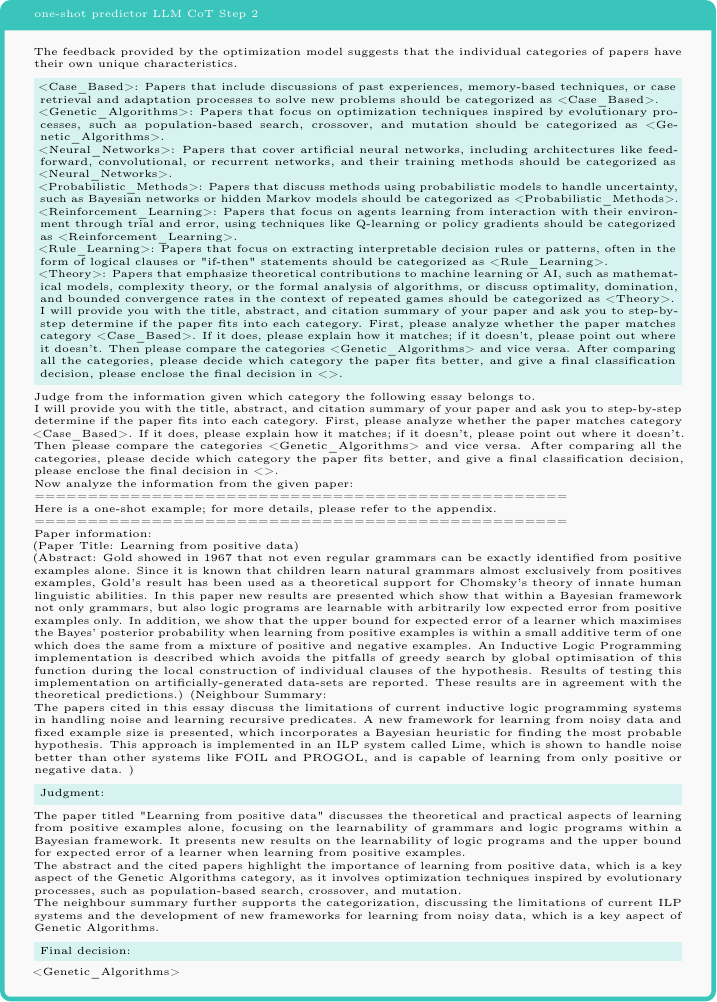}
        \label{fig:one-shot-pred-wp-1}
        \centering
        \includegraphics[scale=0.5]{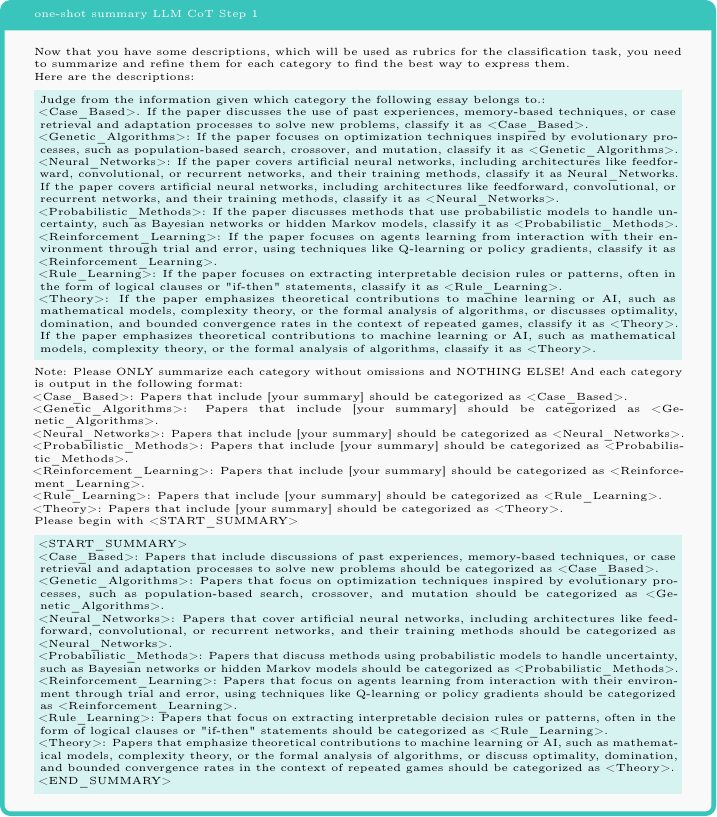}
        \label{fig:one-shot-pred-wp-1}
    \end{minipage}
    \hspace{0.5cm}
    \begin{minipage}{0.6\textwidth}
        \centering
        \includegraphics[scale=0.65]{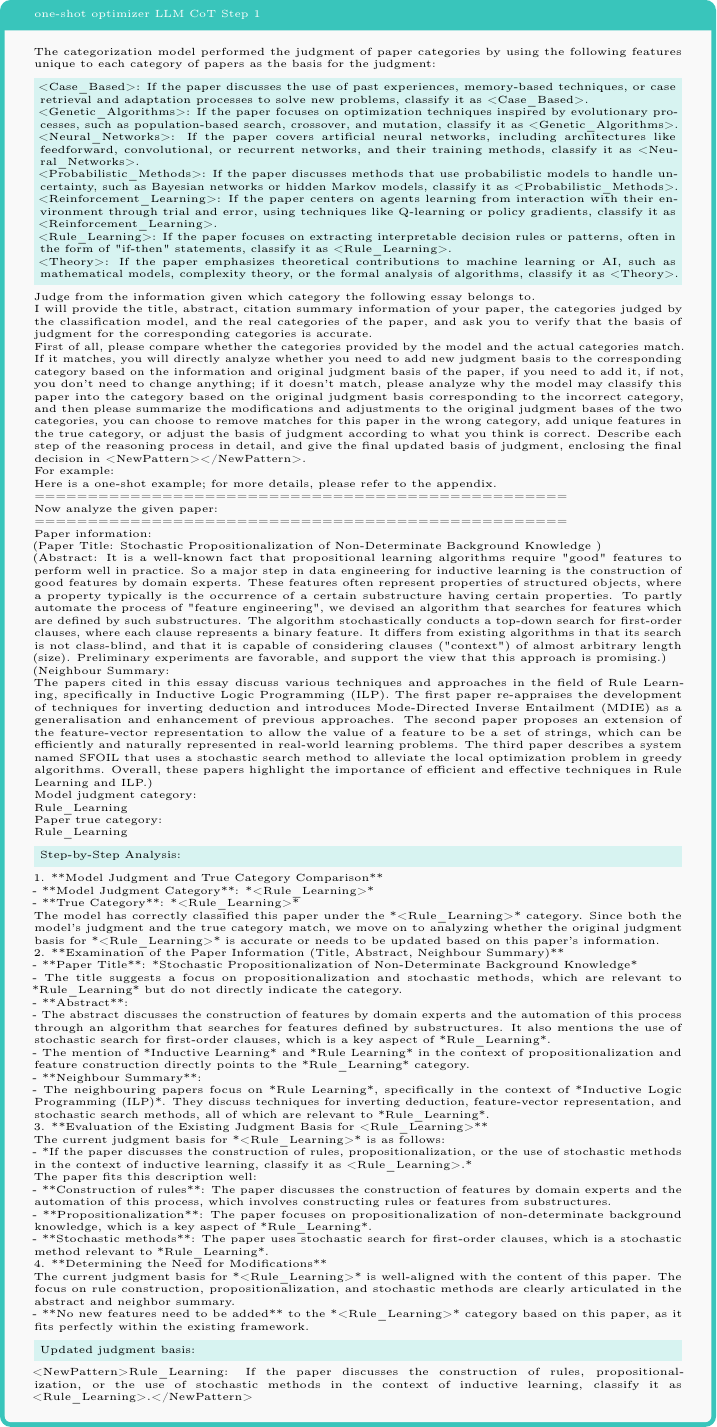}
        \label{fig:one-shot-pred-wp-1}
    \end{minipage}
\end{figure}

\newpage
\subsection{one-shot w/ prior Summary + VGRL Step 2}
\begin{figure}[h]
    \begin{minipage}{0.4\textwidth}
        \centering
        \includegraphics[scale=0.5]{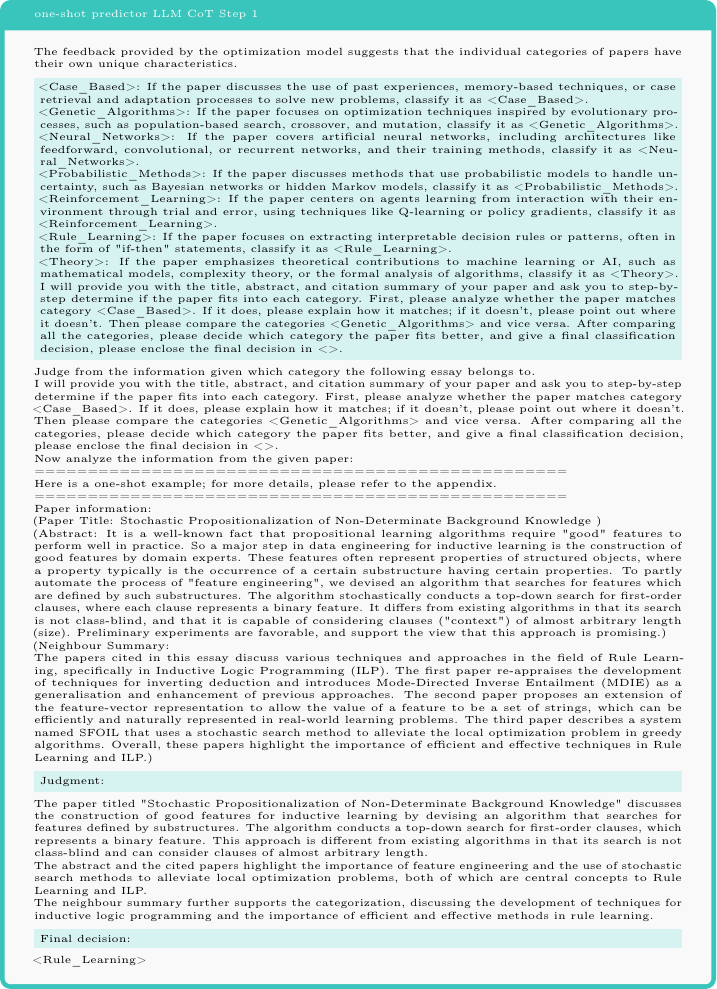}
        \label{fig:one-shot-pred-wp-2}
        \centering
        \includegraphics[scale=0.5]{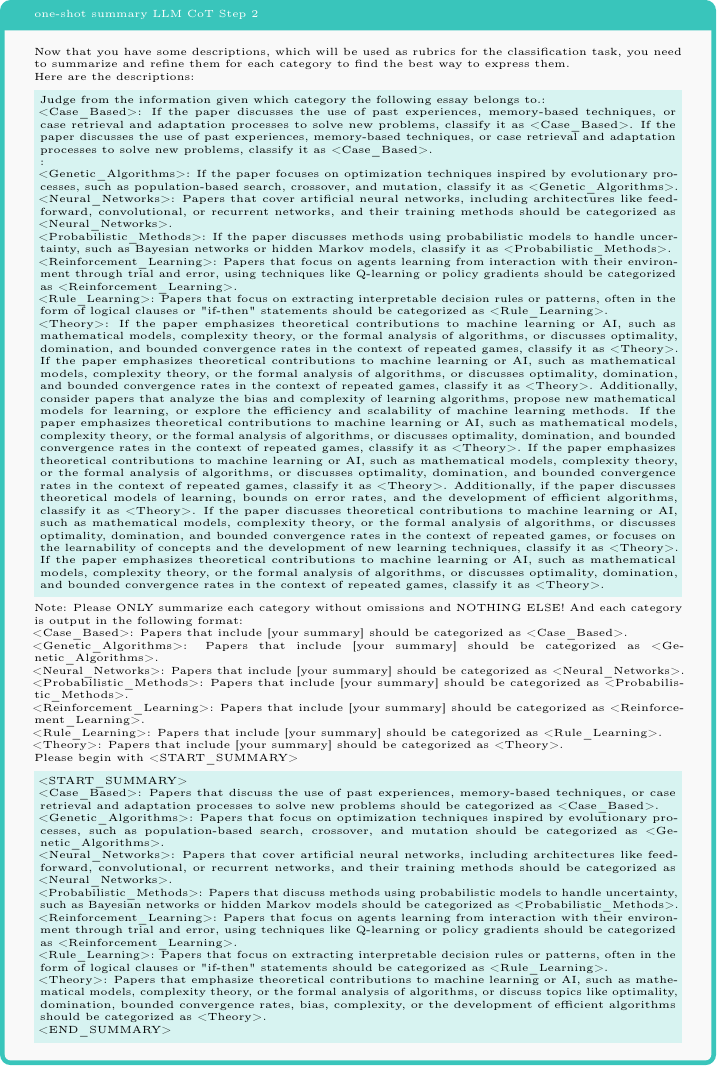}
        \label{fig:one-shot-pred-wp-2}
    \end{minipage}
    \hspace{0.5cm}
    \begin{minipage}{0.6\textwidth}
        \centering
        \includegraphics[scale=0.65]{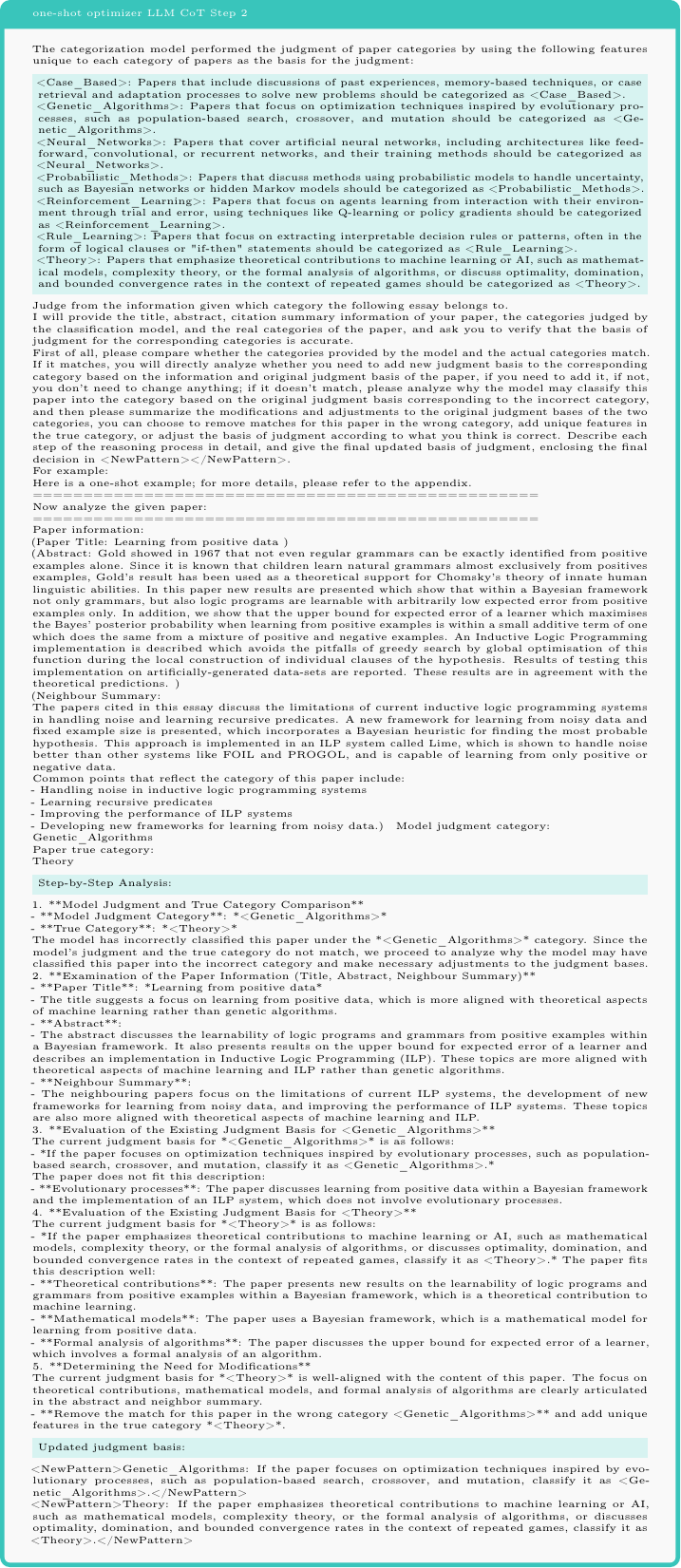}
        \label{fig:one-shot-pred-wp-2}
    \end{minipage}
\end{figure}

\newpage
\subsection{one-shot w/ prior Summary + VGRL Step 80}
\begin{figure}[h]
    \begin{minipage}{0.4\textwidth}
        \centering
        \includegraphics[scale=0.45]{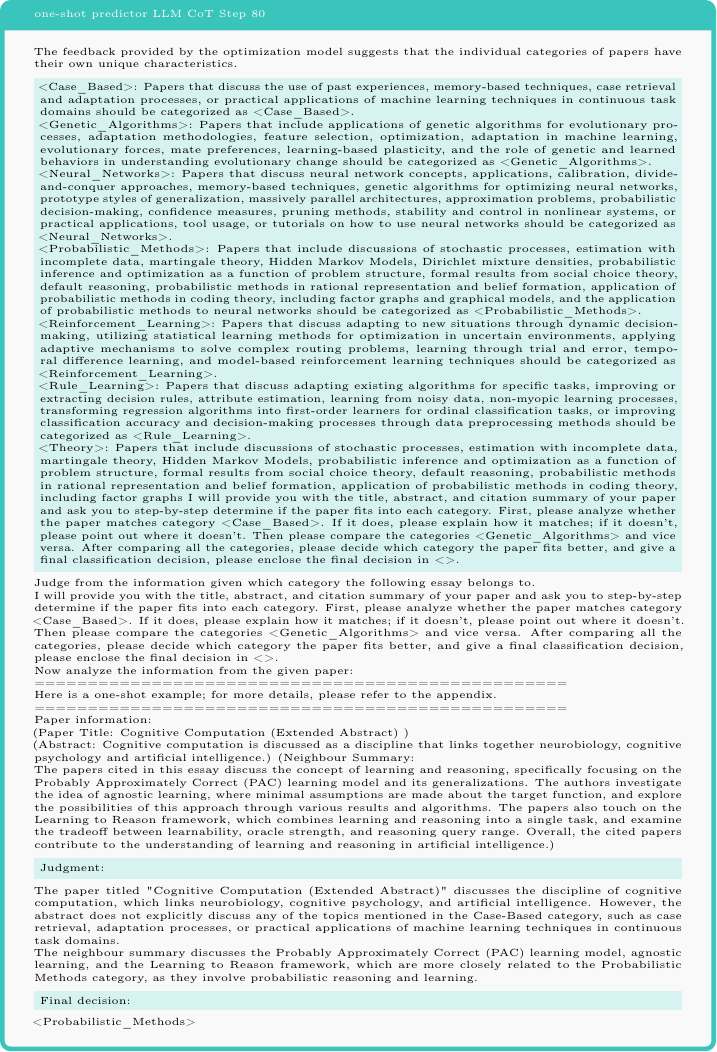}
        \label{fig:one-shot-pred-wp-80}
        \centering
        \includegraphics[scale=0.45]{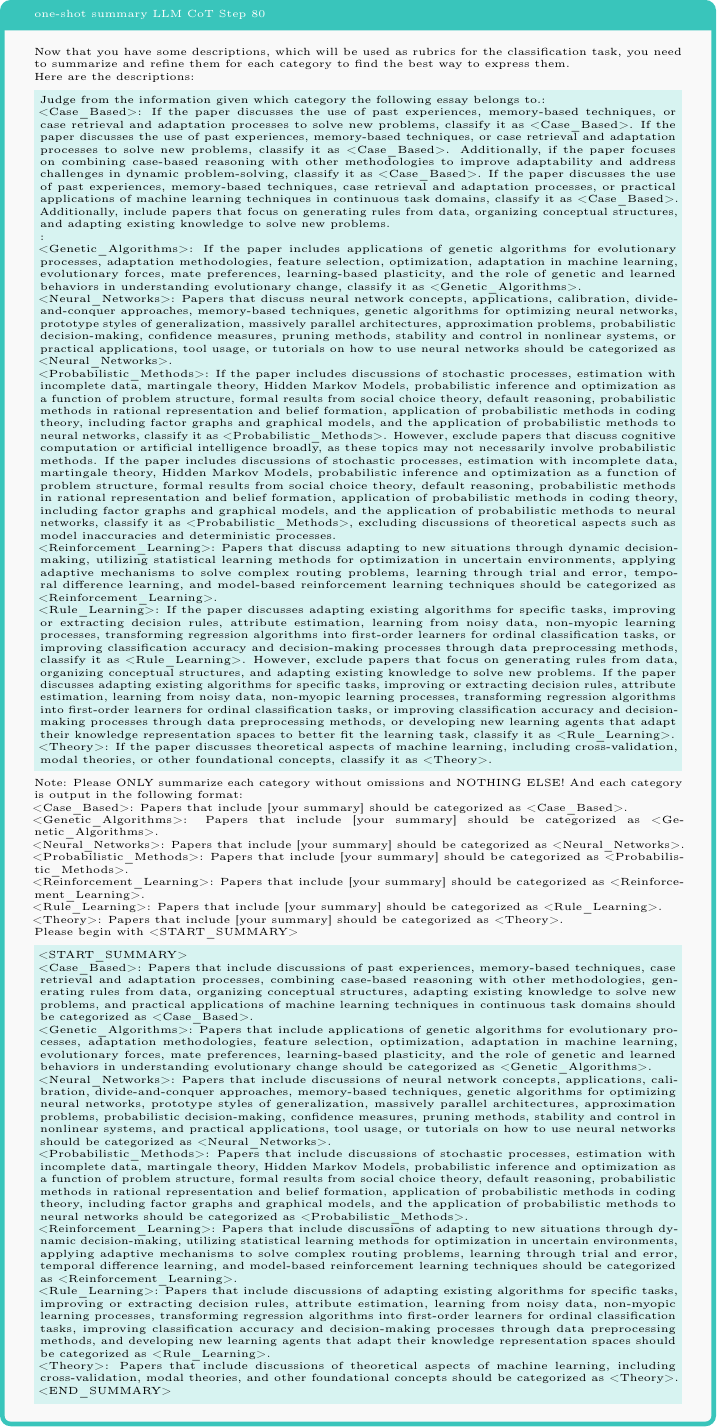}
        \label{fig:one-shot-pred-wp-80}
    \end{minipage}
    \begin{minipage}{0.6\textwidth}
        \centering
        \includegraphics[scale=0.65]{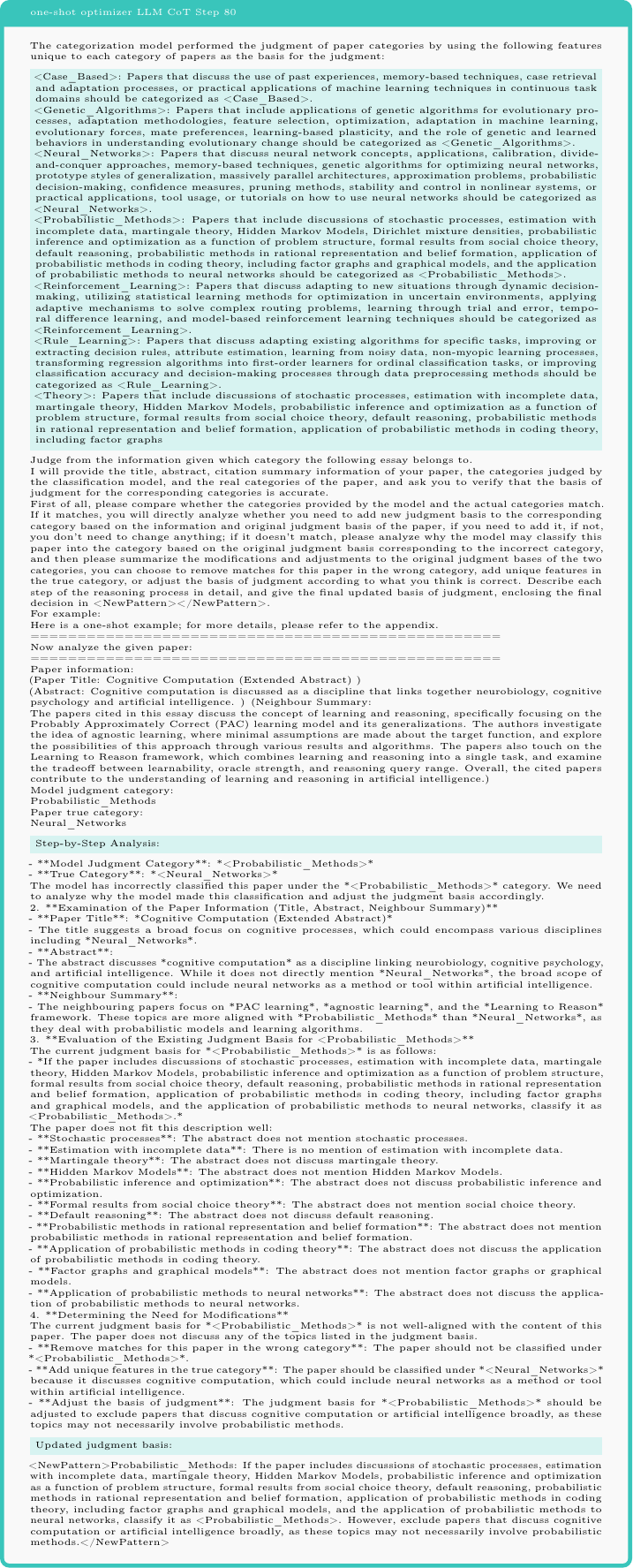}
        \label{fig:one-shot-pred-wp-80}
    \end{minipage}
\end{figure}

\newpage
\section{zero-shot w/o prior Summary + VGRL}\label{F}
\subsection{zero-shot w/o prior Summary + VGRL Step 1}
\begin{figure}[h]
    \begin{minipage}{0.4\textwidth}
        \centering
        \includegraphics[scale=0.5]{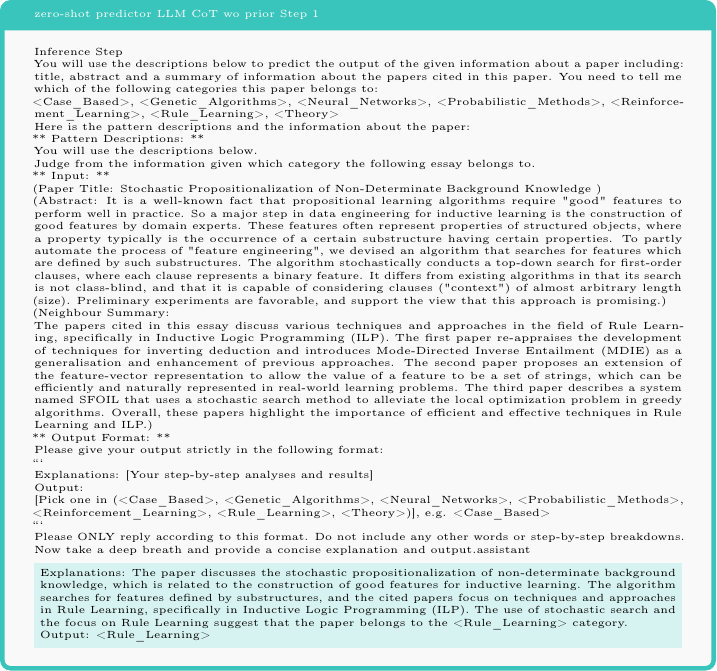}
        \label{fig:zero-shot-pred-wop-1}
        \centering
        \includegraphics[scale=0.5]{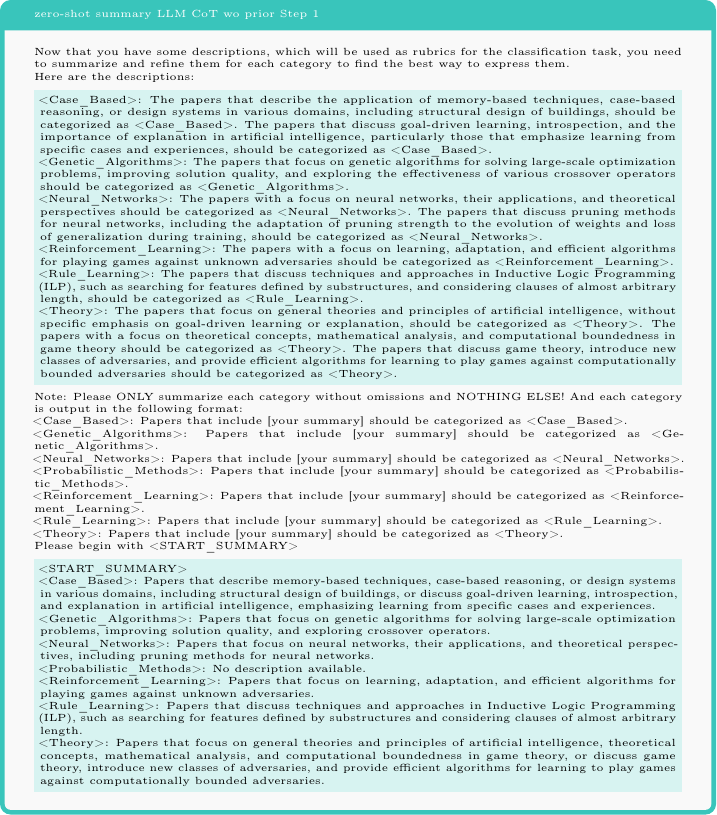}
        \label{fig:zero-shot-pred-wop-1}
    \end{minipage}
    \hspace{0.5cm}
    \begin{minipage}{0.6\textwidth}
        \centering
        \includegraphics[scale=0.8]{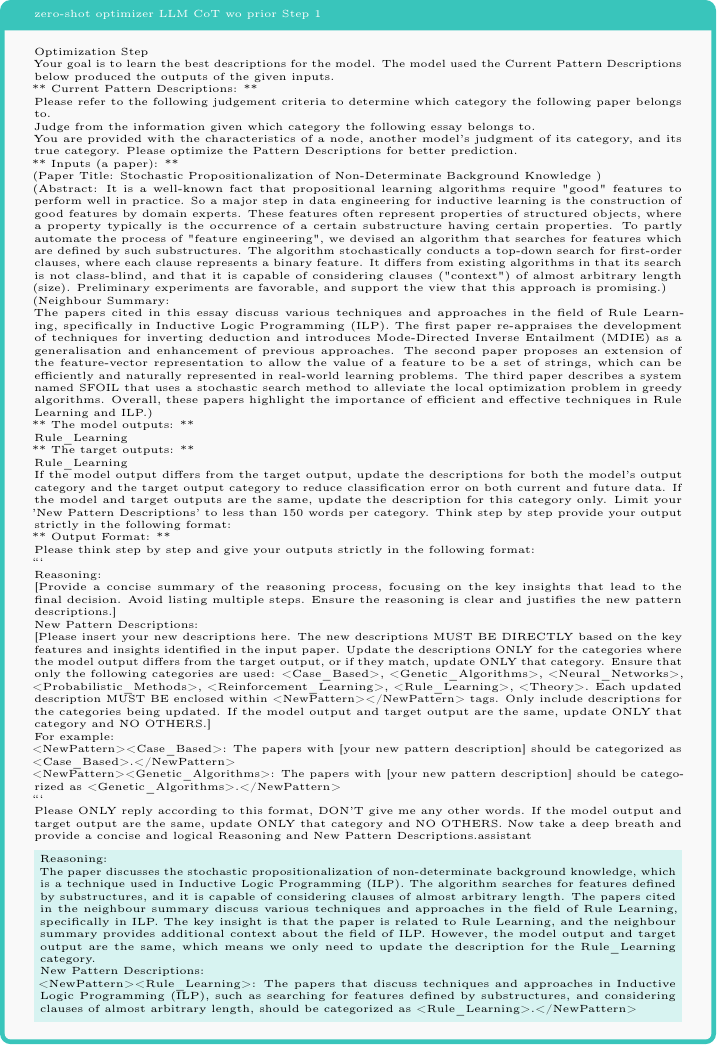}
        \label{fig:zero-shot-pred-wop-1}
    \end{minipage}
\end{figure}

\newpage
\subsection{zero-shot w/o prior Summary + VGRL Step 2}
\begin{figure}[h]
    \begin{minipage}{0.4\textwidth}
        \centering
        \includegraphics[scale=0.5]{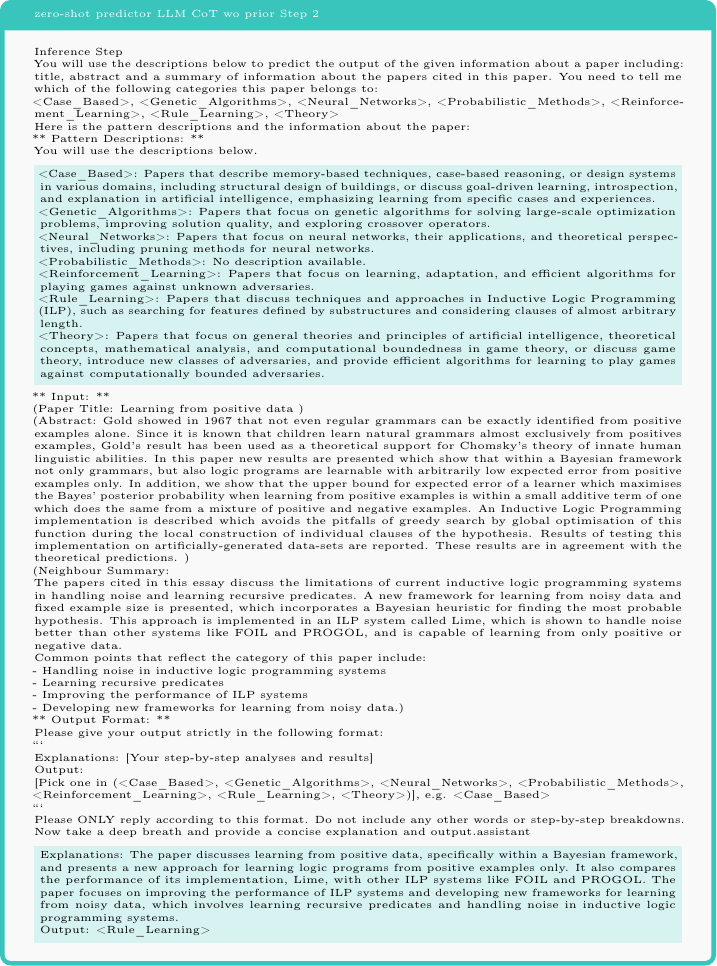}
        \label{fig:zero-shot-pred-wop-2}
        \centering
        \includegraphics[scale=0.5]{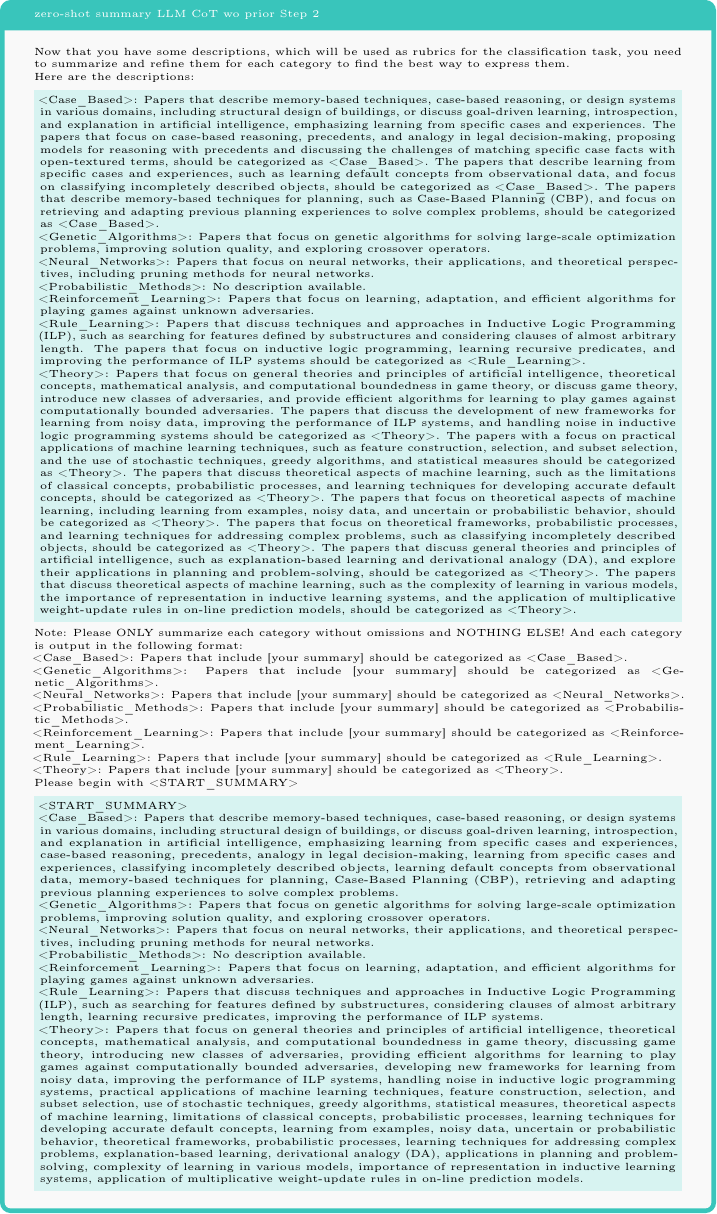}
        \label{fig:zero-shot-pred-wop-2}
    \end{minipage}
    \hspace{0.5cm}
    \begin{minipage}{0.6\textwidth}
        \centering
        \includegraphics[scale=0.8]{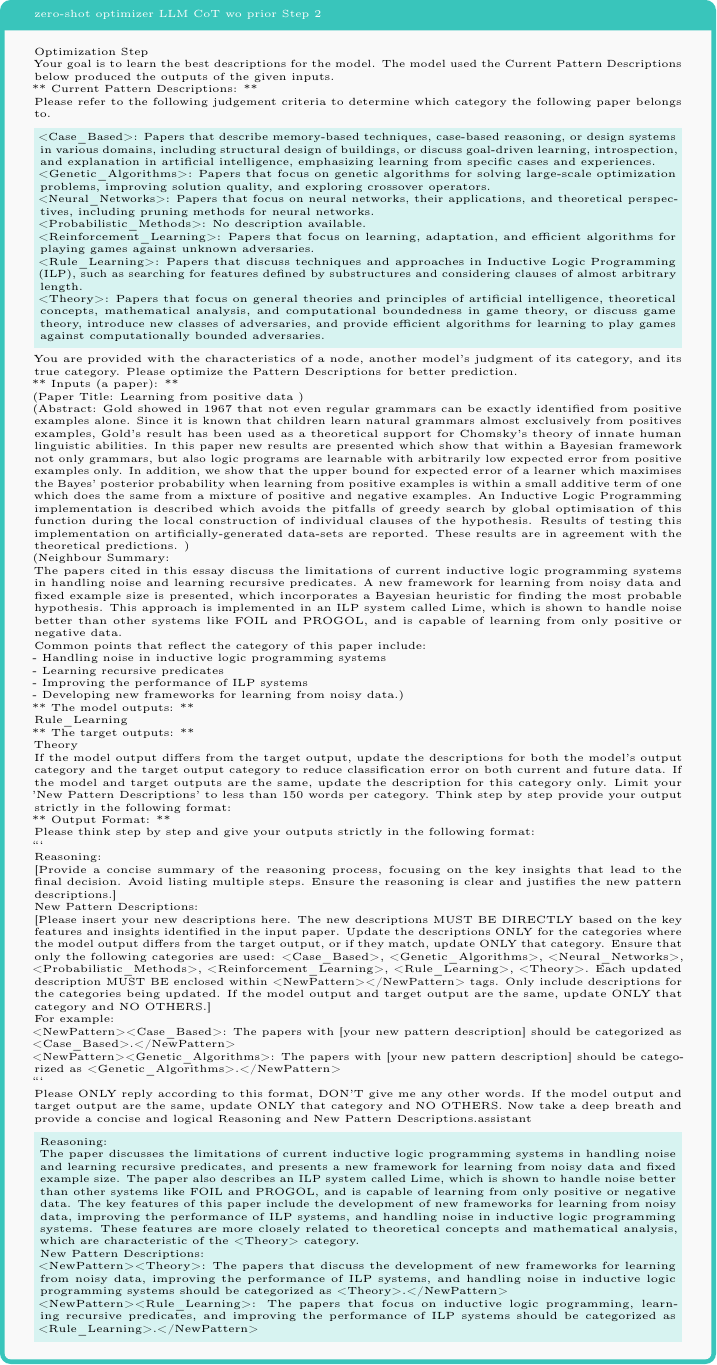}
        \label{fig:zero-shot-pred-wop-2}
    \end{minipage}
\end{figure}

\newpage
\subsection{zero-shot w/o prior Summary + VGRL Step 80}
\begin{figure}[h]
    \begin{minipage}{0.4\textwidth}
        \centering
        \includegraphics[scale=0.5]{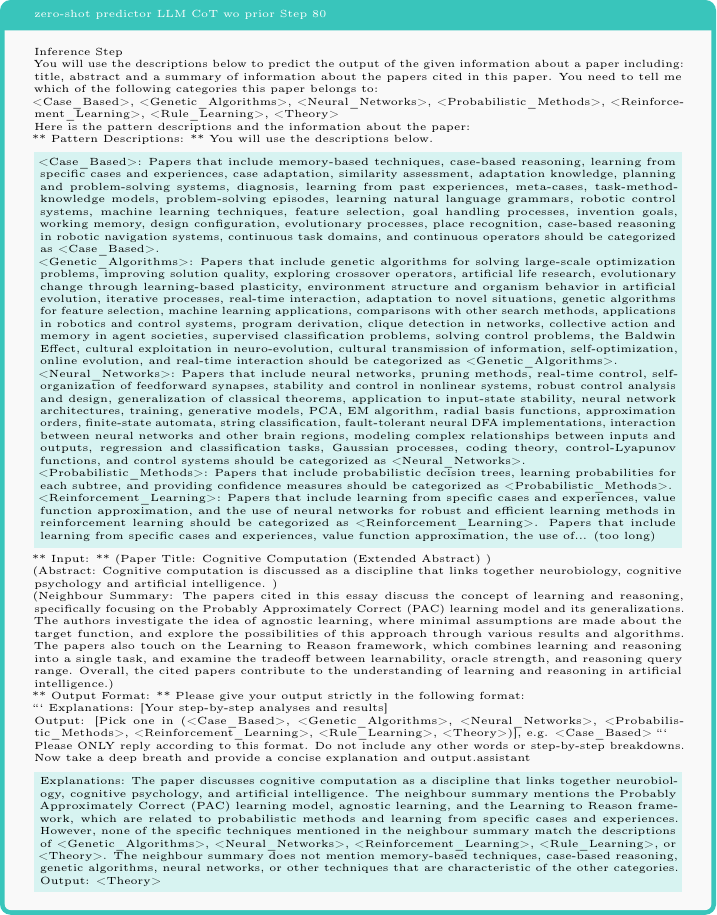}
        \label{fig:zero-shot-pred-wop-80}
        \centering
        \includegraphics[scale=0.5]{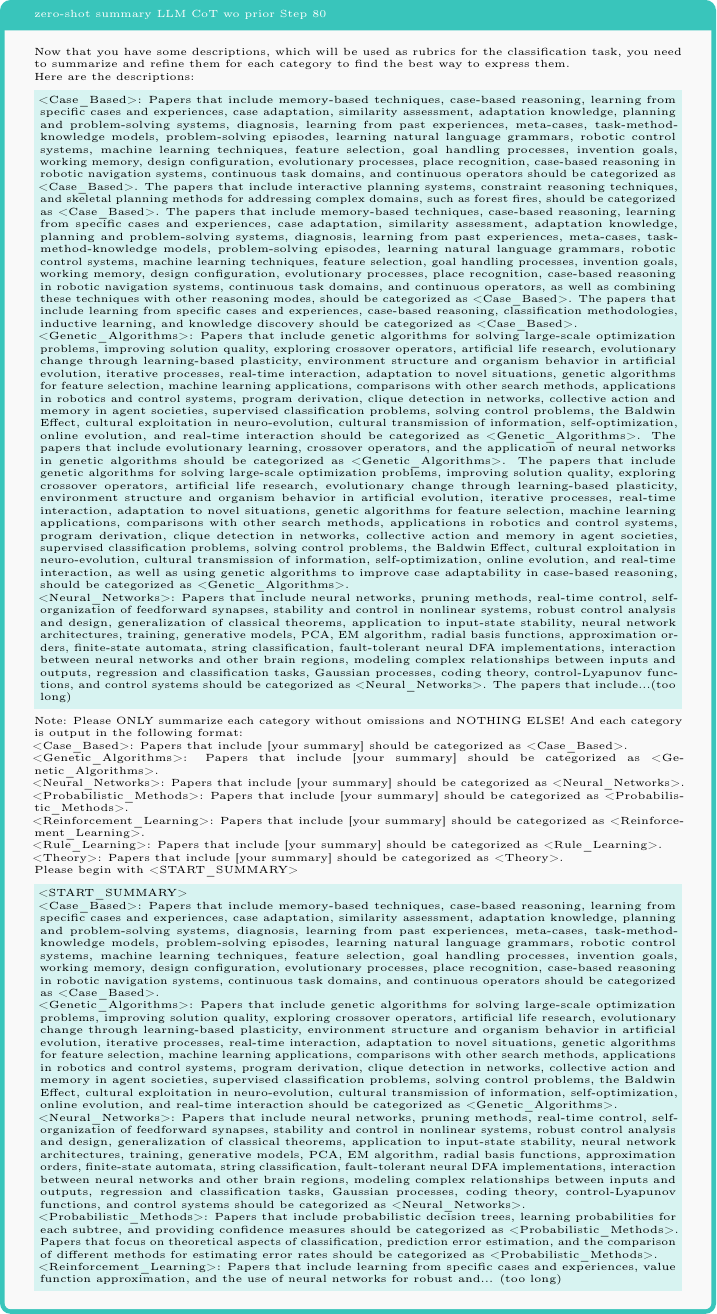}
        \label{fig:zero-shot-pred-wop-80}
    \end{minipage}
    \hspace{0.5cm}
    \begin{minipage}{0.6\textwidth}
        \centering
        \includegraphics[scale=0.8]{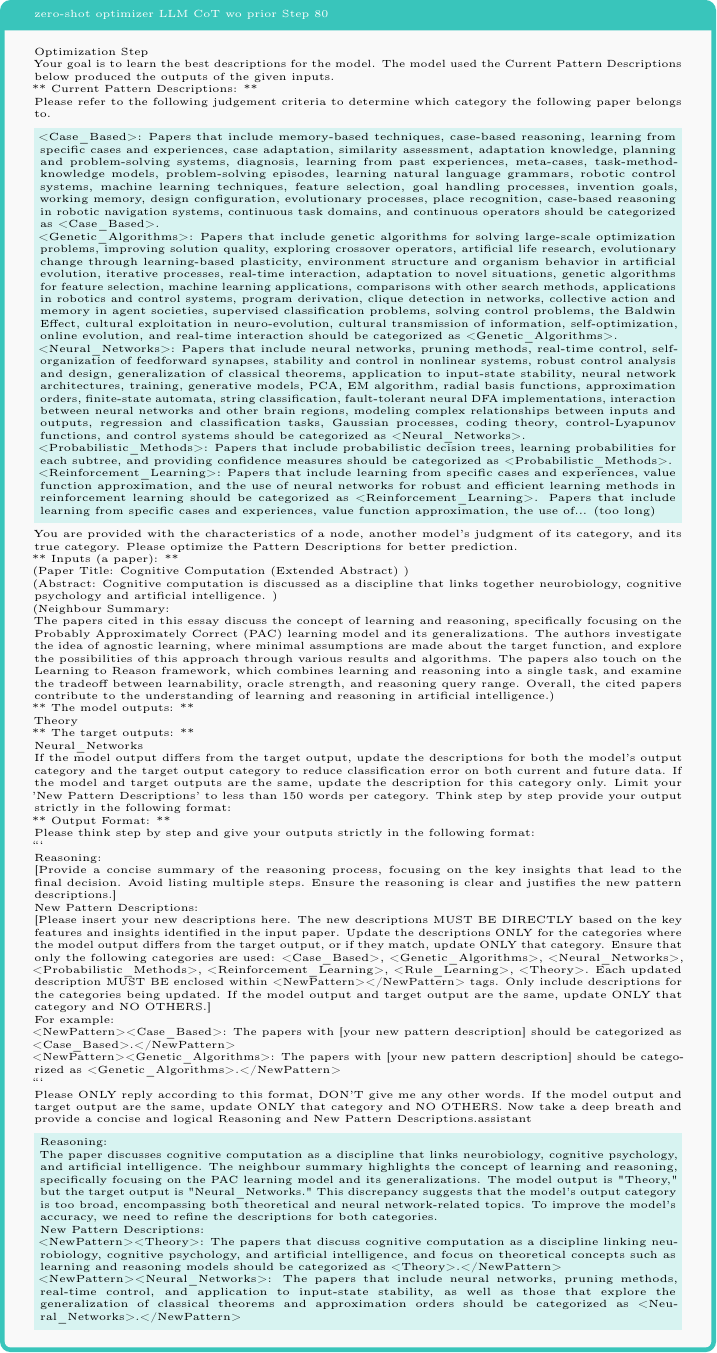}
        \label{fig:zero-shot-pred-wop-80}
    \end{minipage}
\end{figure}

\newpage
\section{zero-shot w/ prior Summary + VGRL}\label{G}
\subsection{zero-shot w/ prior Summary + VGRL Step 1}
\begin{figure}[h]
    \begin{minipage}{0.4\textwidth}
        \centering
        \includegraphics[scale=0.5]{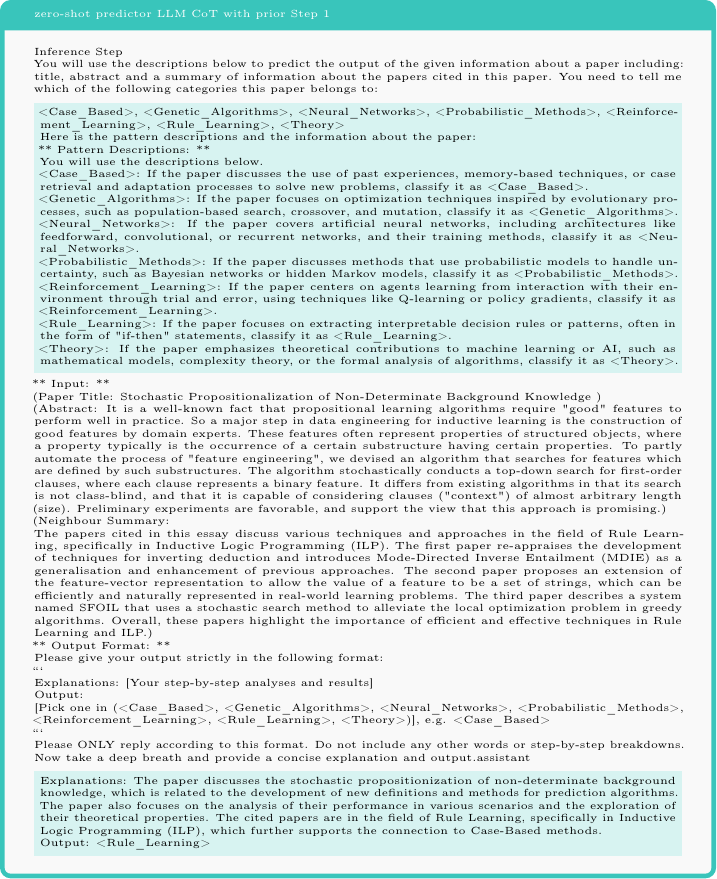}
        \label{fig:zero-shot-pred-wp-1}
        \centering
        \includegraphics[scale=0.5]{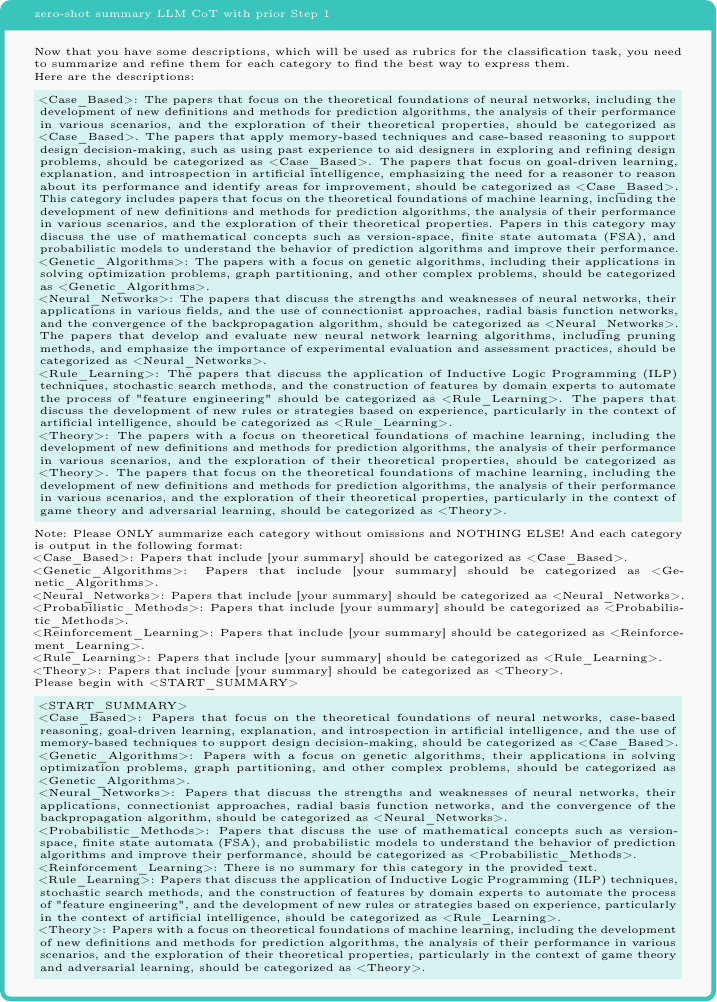}
        \label{fig:zero-shot-pred-wp-1}
    \end{minipage}
    \hspace{0.5cm}
    \begin{minipage}{0.6\textwidth}
        \centering
        \includegraphics[scale=0.8]{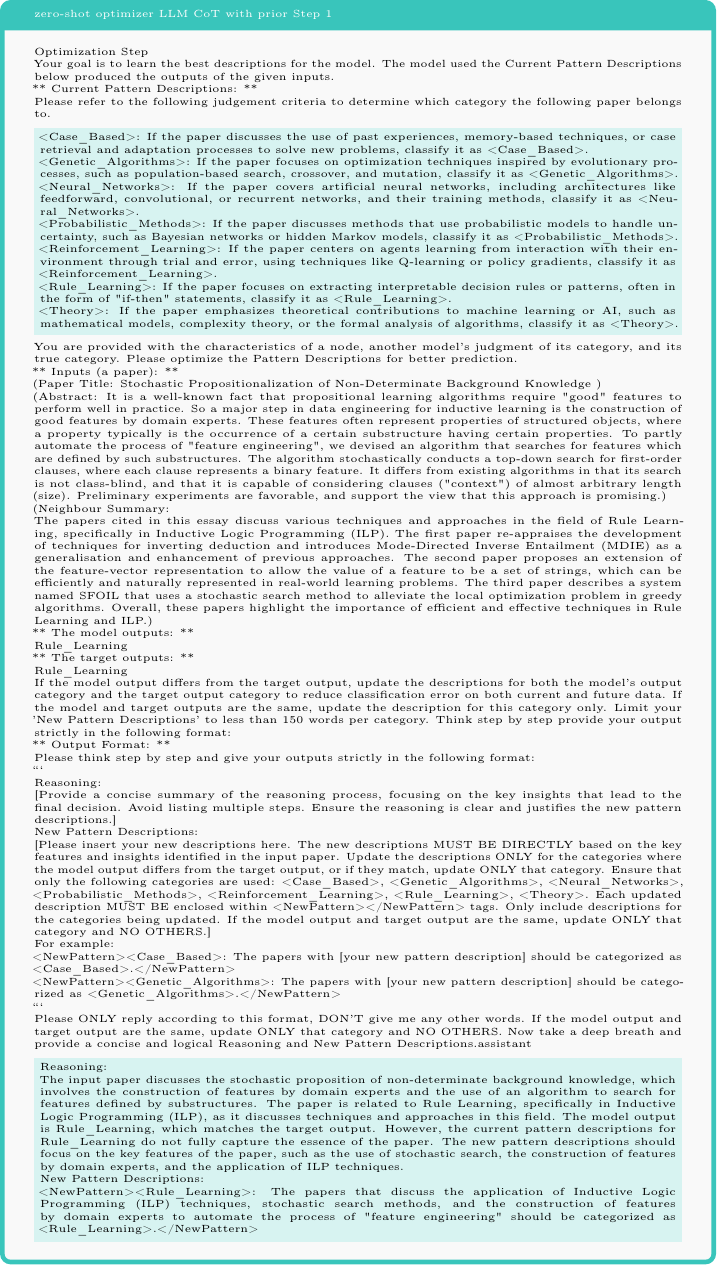}
        \label{fig:zero-shot-pred-wp-1}
    \end{minipage}
\end{figure}

\newpage
\subsection{zero-shot w/ prior Summary + VGRL Step 2}
\begin{figure}[h]
    \begin{minipage}{0.4\textwidth}
        \centering
        \includegraphics[scale=0.5]{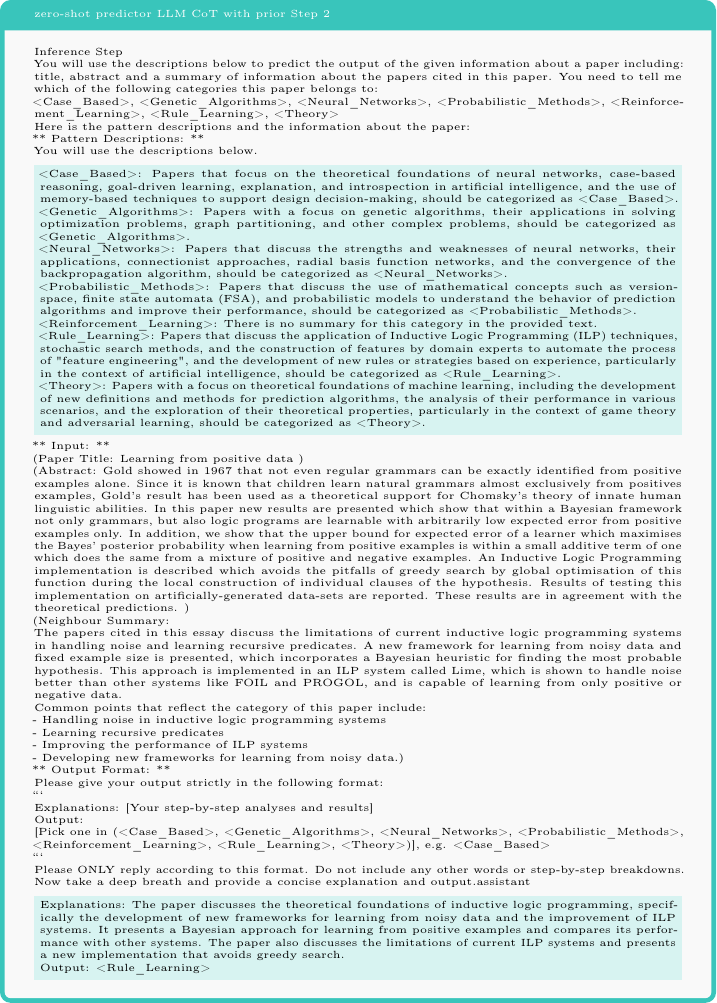}
        \label{fig:zero-shot-pred-wp-2}
        \centering
        \includegraphics[scale=0.5]{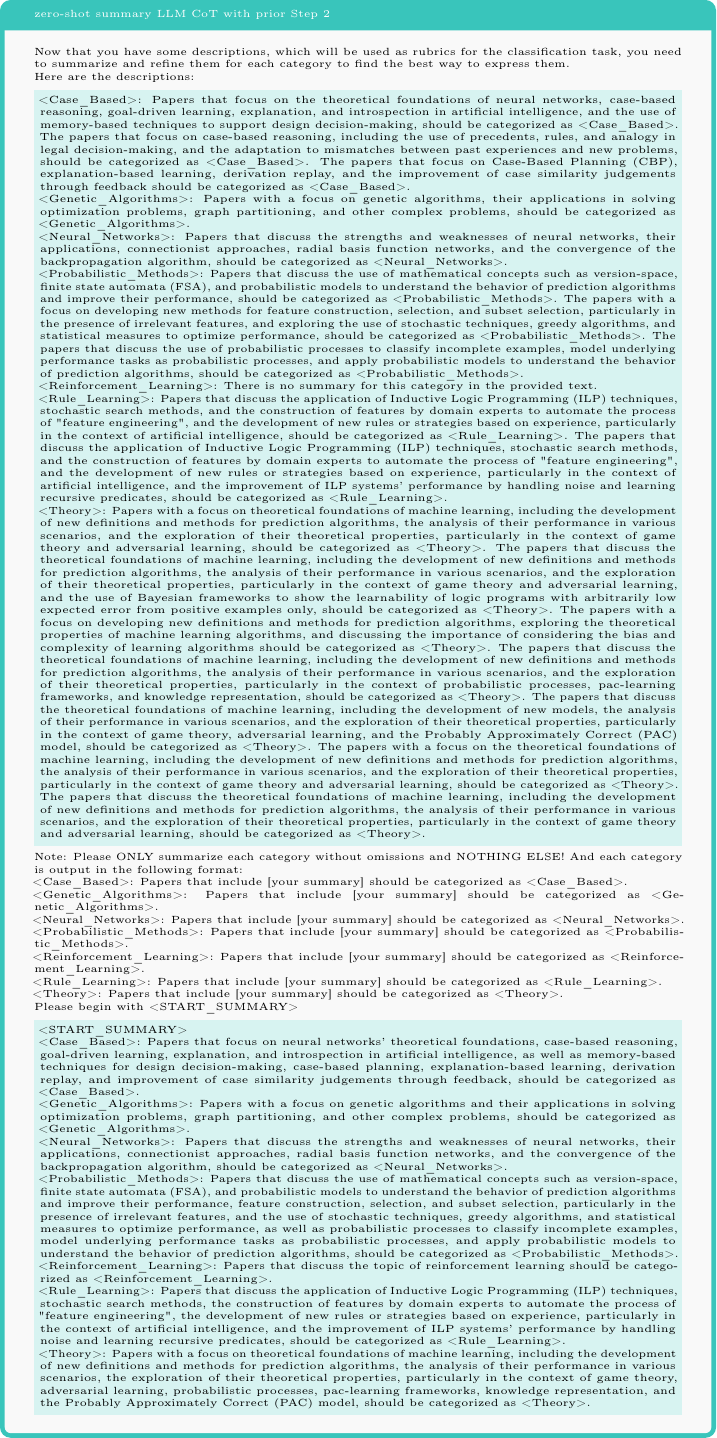}
        \label{fig:zero-shot-pred-wp-2}
    \end{minipage}
    \hspace{0.5cm}
    \begin{minipage}{0.6\textwidth}
        \centering
        \includegraphics[scale=0.8]{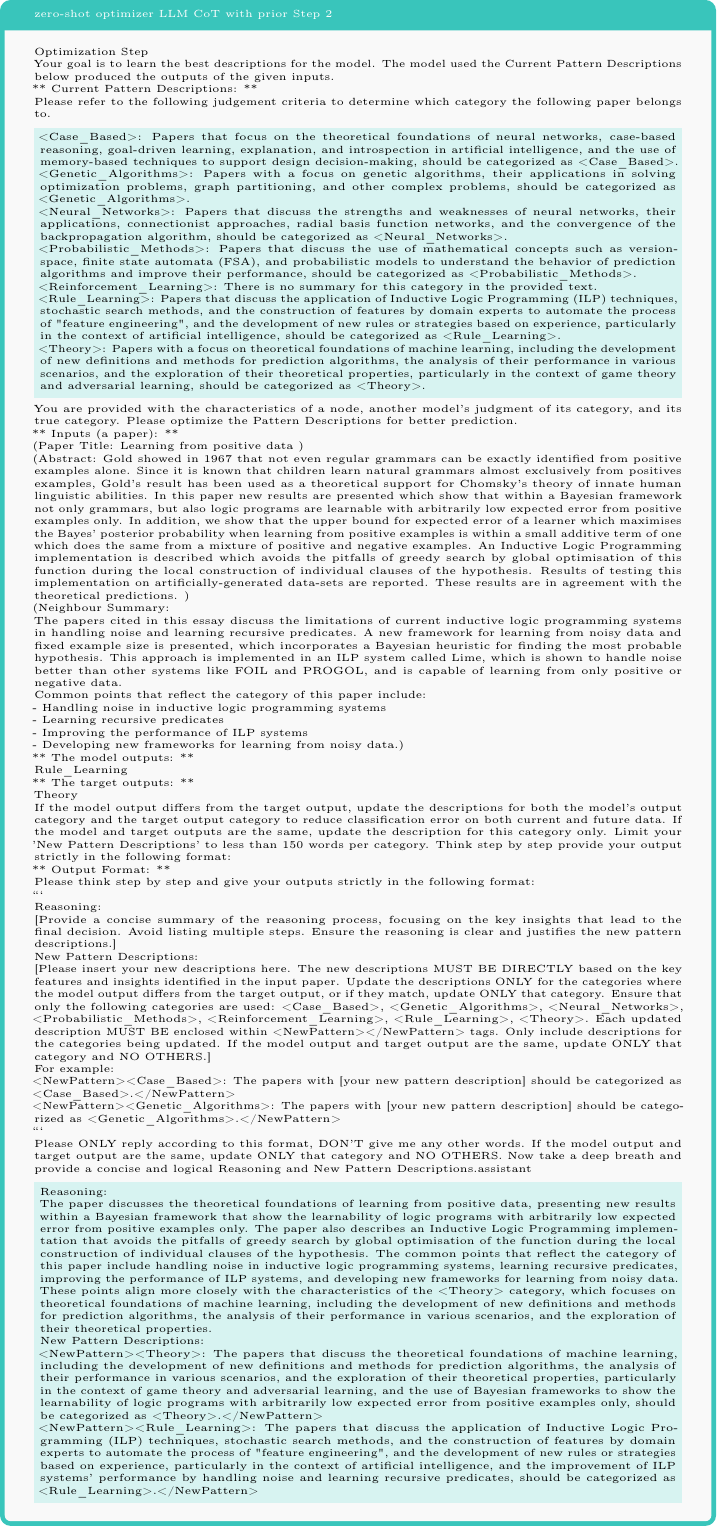}
        \label{fig:zero-shot-pred-wp-2}
    \end{minipage}
\end{figure}

\newpage
\subsection{zero-shot w/ prior Summary + VGRL Step 80}
\begin{figure}[h]
    \begin{minipage}{0.4\textwidth}
        \centering
        \includegraphics[scale=0.5]{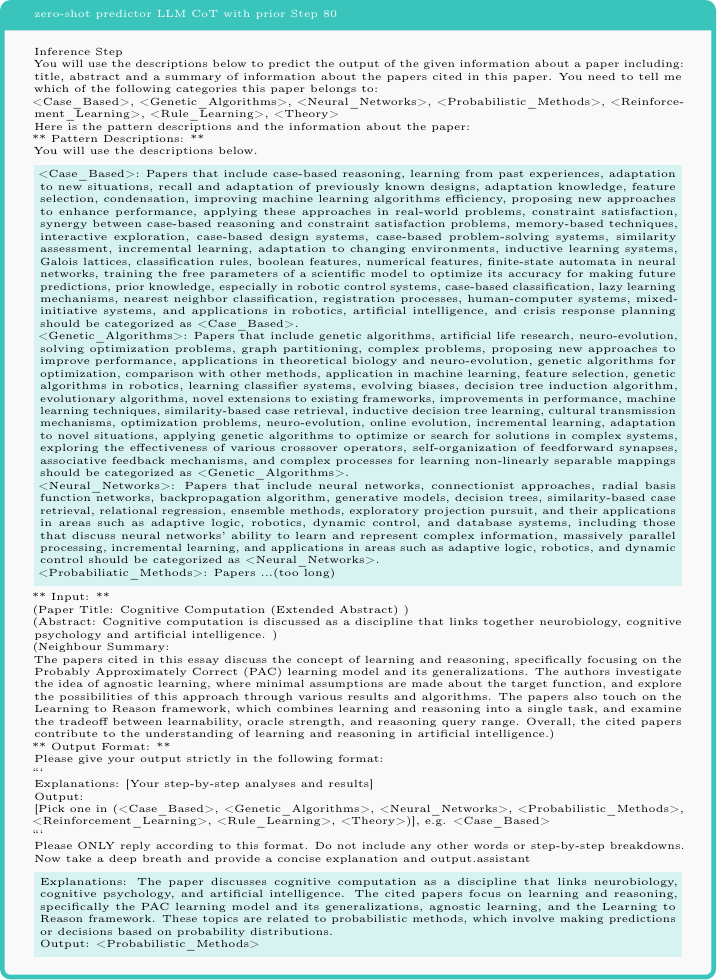}
        \label{fig:zero-shot-pred-wp-80}
        \centering
        \includegraphics[scale=0.5]{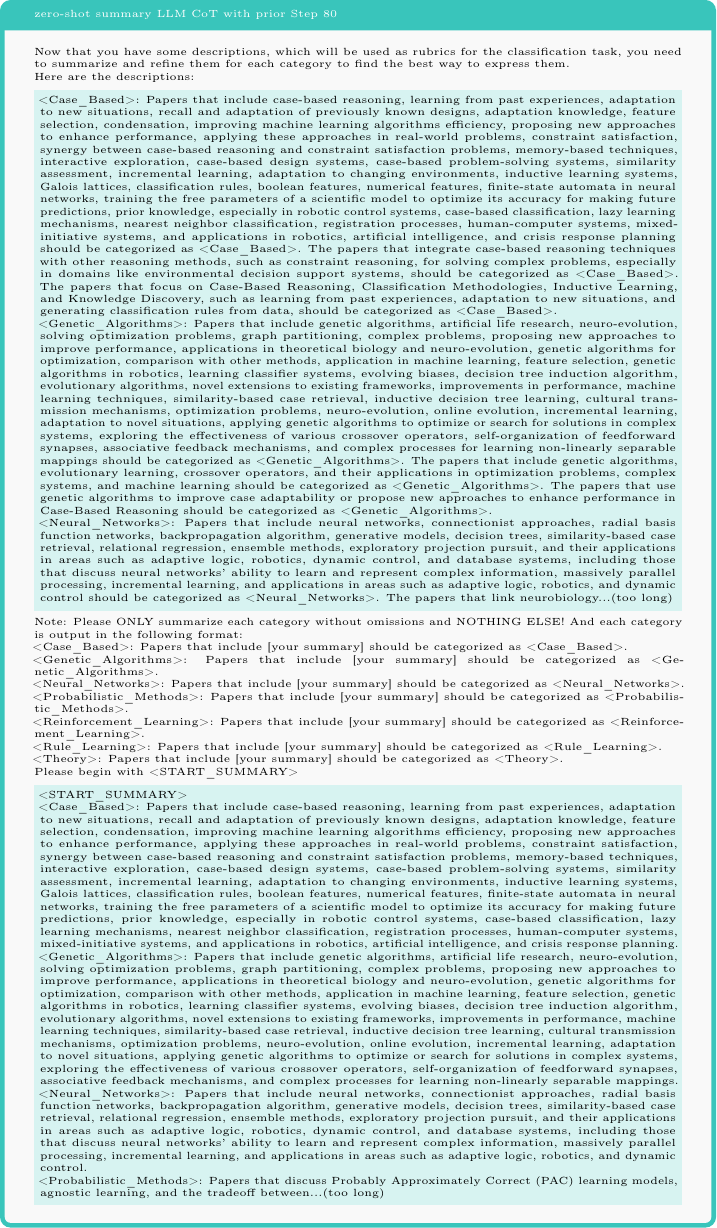}
        \label{fig:zero-shot-pred-wp-80}
    \end{minipage}
    \hspace{0.5cm}
    \begin{minipage}{0.6\textwidth}
        \centering
        \includegraphics[scale=0.8]{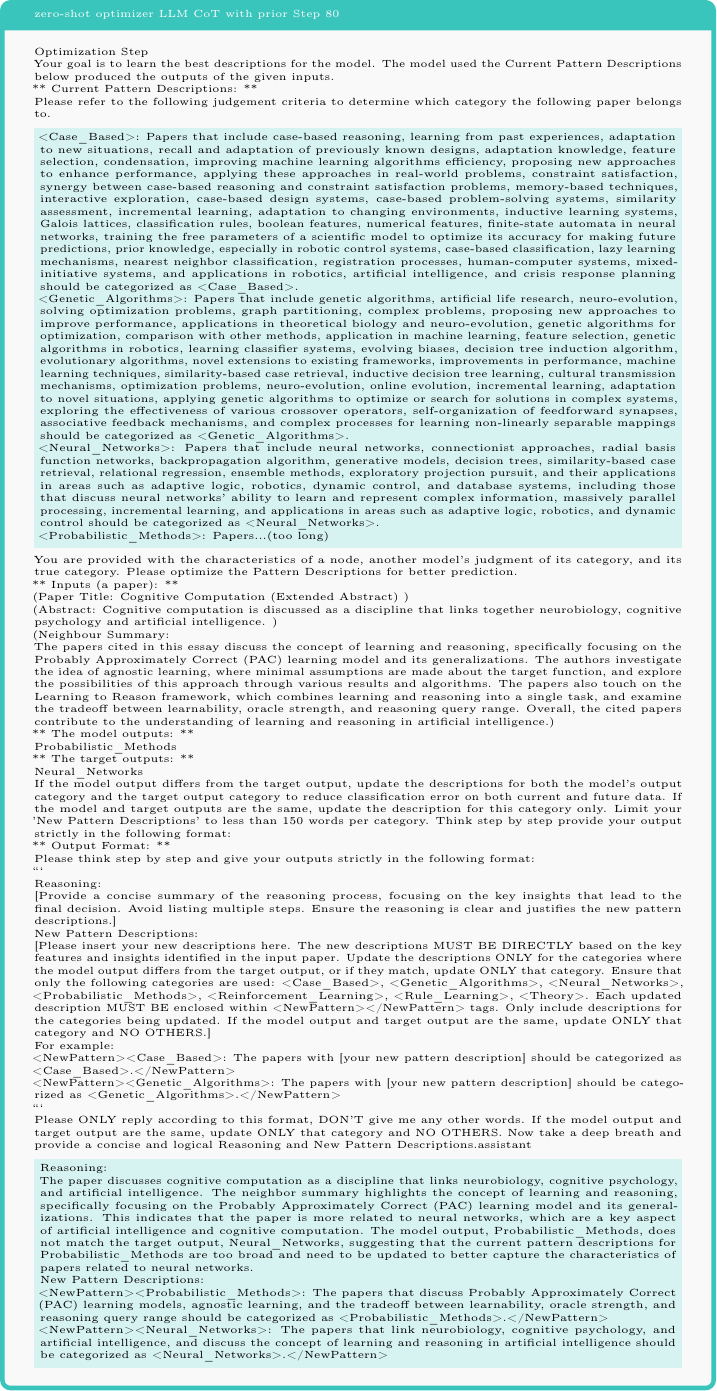}
        \label{fig:zero-shot-pred-wp-80}
    \end{minipage}
\end{figure}

\end{document}